\documentclass[acmlarge]{acmart}
\AtBeginDocument{%
  \providecommand\BibTeX{{%
    \normalfont B\kern-0.5em{\scshape i\kern-0.25em b}\kern-0.8em\TeX}}}

\setcopyright{acmlicensed}
\copyrightyear{2024}
\acmYear{2024}
\acmDOI{XXXXXXX.XXXXXXX}

\usepackage{booktabs}
\usepackage{makecell}
\usepackage{hyperref}
\usepackage{xurl}
\usepackage{graphicx}
\usepackage{threeparttable}
\usepackage{threeparttablex}
\usepackage{tikz}
\usepackage{xcolor}
\usepackage{multirow}

\definecolor{purple}{RGB}{99, 88, 232}

\newcommand*\circled[3]{\tikz[baseline=(char.base)]{
            \node[shape=circle,draw=#1,fill=#1,inner sep=0.3pt,text=#2] (char) {#3};}}

\begin{document}

\title{HRDE: Retrieval-Augmented Large Language Models for Chinese Health Rumor Detection and Explainability}

\author{Yanfang Chen}
\authornote{Both authors contributed equally to this research.}
\email{cyf@ruc.edu.cn}
\affiliation{%
  \institution{Renmin University of China}
  \city{Beijing}
  \country{China}
}

\author{Ding Chen}
\authornotemark[1]
\affiliation{%
  \institution{Institute for Advanced Algorithms Research, Shanghai}
  \city{Shanghai}
  \country{China}
}

\author{Shichao Song}
\author{Simin Niu}
\author{Hanyu Wang}
\affiliation{%
  \institution{Renmin University of China}
  \city{Beijing}
  \country{China}
}

\author{Zeyun Tang}
\author{Feiyu Xiong}
\author{Zhiyu Li}
\authornote{Corresponding author.}
\affiliation{%
  \institution{Institute for Advanced Algorithms Research, Shanghai}
  \city{Shanghai}
  \country{China}
}

\renewcommand{\shortauthors}{Trovato and Tobin, et al.}

\begin{abstract}
  As people increasingly prioritize their health, the speed and breadth of health information dissemination on the internet have also grown. At the same time, the presence of false health information (health rumors) intermingled with genuine content poses a significant potential threat to public health. However, current research on Chinese health rumors still lacks a large-scale, public, and open-source dataset of health rumor information, as well as effective and reliable rumor detection methods. This paper addresses this gap by constructing a dataset containing 1.12 million health-related rumors (HealthRCN) through web scraping of common health-related questions and a series of data processing steps. HealthRCN is the largest known dataset of Chinese health information rumors to date. Based on this dataset, we propose retrieval-augmented large language models for Chinese health rumor detection and explainability (HRDE \footnote{HRDE is currently deployed at \url{http://www.rumors.icu/}.} \footnote{The source code is available at GitHub: \url{https://github.com/hush-cd/HRDE}}). This model leverages retrieved relevant information to accurately determine whether the input health information is a rumor and provides explanatory responses, effectively aiding users in verifying the authenticity of health information. In evaluation experiments, we compared multiple models and found that HRDE outperformed them all, including GPT-4-1106-Preview, in rumor detection accuracy and answer quality. HRDE achieved an average accuracy of 91.04\% and an F1 score of 91.58\%. 
\end{abstract}




\maketitle

\section{Introduction}
With the development of socio-economic conditions and the improvement of living standards, health has gradually garnered people's attention, becoming a crucial aspect of modern people's pursuit of a high-quality life~\cite{28,29}. At the same time, in the face of emerging health challenges, such as the increase in global pandemics and chronic diseases, public health awareness has been further stimulated, leading to a surge in the demand for reliable online health information. However, the grassroots and anonymous nature of the internet has directly or indirectly resulted in an explosive growth of health rumors in recent years. These health rumors may include unverified treatments, unscientific health advice, and even incorrect medical information, all of which can have serious negative impacts on public health~\cite{30}. Therefore, there is an urgent need for effective methods to detect health rumors in online media.

In the current landscape of health rumor detection research, one of the primary limiting factors is the lack of publicly available health rumor corpora, particularly those involving long-text health rumors. Most existing studies rely on self-collected data for identification methods and predominantly focus on short-text datasets such as those from Twitter ~\cite{1}. However, health rumors differ from general rumors in that they are not merely expressions of opinions. They often consist of mid-to-long narratives that interweave rumor information with factual content, enhancing their spreadability. This characteristic contributes significantly to the popularity of health rumors on platforms like WeChat\footnote{WeChat is a Chinese multi-purpose messaging, social media, and mobile payment app developed by Tencent. It is widely used for instant messaging, voice and video calls, social networking, and various services like news, games, and financial transactions.}. Consequently, there has been increasing attention on the detection of long-text health rumors~\cite{31}. 

Another critical shortcoming in current health rumor detection methods is the lack of interpretability in the results. Given the informative and scientific nature of health information, it is essential that the detection results extend beyond a simple binary classification of "true" or "false." Similar to manual refuting, the detection outcomes should provide comprehensive explanations and authoritative sources to gain users' trust.

In recent years, the emergence and development of large language models (LLMs) have presented new opportunities for applications across various industries. The application of several LLMs in the health domain has also been proposed ~\cite{2}. For example, Bao et al. proposed DISC-MedLLM, which utilizes LLMs to achieve end-to-end conversational medical services~\cite{32}. However, it is important to note that while these models offer exceptional natural language processing and comprehension capabilities, they also introduce certain hallucination issues ~\cite{3}. Considering the technical capabilities of LLMs in handling health information, and the specific requirements for accuracy and reliability in the health information field, we face a challenge: although LLMs excel in natural language understanding, they may introduce misleading information when generating content, known as the "hallucination" problem. This necessitates more precise adjustments and optimizations in the detection and explainability of health-related rumors.

Concretely, we constructed a Chinese rumor dataset containing over 1.12 million data points, named Health Rumor CN (HealthRCN), to fill the gap of Chinese rumor datasets in the health domain. Simultaneously, we propose a novel model called Retrieval-Augmented Large Language Models for Chinese \textbf{H}ealth \textbf{R}umor \textbf{D}etection and \textbf{E}xplainability (HRDE). This model comprises four components: health information collection and storage, health information retrieval, retrieval information re-ranking, and the LLM generating the rumor detection answer (including rumor detection and analysis). The main contributions of this work are summarized as follows:
\begin{itemize}
  \item We have constructed the Health Rumor CN (HealthRCN) dataset, filling the gap in Chinese rumor datasets in the health information domain.
  \item We proposed a unified framework for rumor detection that enables real-time access to the latest health information and performs rumor detection on user-input health information, providing an interpretable analysis process.
  \item We evaluated HRDE and multiple models on the HealthRCN dataset, surpassing all models in terms of rumor detection accuracy and answer quality metrics.
\end{itemize}

The structure of the remaining sections of this paper is as follows: Section ~\ref{sec:Related Works} examines the related works; Section ~\ref{sec:HRDE} provides a detailed description of the HRDE framework and its implementation process; Section ~\ref{sec:Experiment} presents the evaluation experiments conducted with HRDE and multiple models, along with the analysis of the experimental results; Section ~\ref{sec: Case Study} analyzes three practical cases of rumor detection and answer using HRDE; Section ~\ref{sec:Conclusion} concludes the paper with a summary.

\section{Related Works}
\label{sec:Related Works}
\subsection{Health Rumor Detection}

\subsubsection{Machine Learning-Based Rumor Detection Methods}

These methods primarily rely on traditional machine learning theories, using manually defined features as model inputs. Such features may include rumor text features~\cite{8,9}, user behavior features~\cite{10,11}, network features~\cite{12}, among others, which are then combined with relevant classification algorithms for screening. For example, Marco L et al. propose a novel ML fake news detection method that combines news content and social context features~\cite{13}. Ma J et al. propose a kernel-based method called Propagation Tree Kernel, which captures high-order patterns differentiating different types of rumors by evaluating the similarities between their propagation tree structures~\cite{14}. The advantage of machine learning-based rumor detection methods lies in their relative simplicity and lower computational complexity. However, they depend on manually designed features, which can somewhat limit the model's generalization capability and adaptability to new types of rumors.

\subsubsection{Deep Learning-Based Rumor Detection Methods}

These methods utilize deep neural networks (such as CNN~\cite{15}, RNN~\cite{16}, LSTM~\cite{17}) to automatically learn complex feature representations from rumor information, thereby improving the accuracy and robustness of rumor detection and overcoming the limitations of manually extracted features. For example, Bian T. et al. propose a bi-directional graph model, named Bi-Directional Graph Convolutional Networks (Bi-GCN), to explore both characteristics by operating on both top-down and bottom-up propagation of rumors~\cite{18}. Liu T. et al. introduce a new detection model that jointly learns both the representations of user correlation and information propagation to detect rumors on social media~\cite{19}. Deep learning-based methods are better at capturing complex features and patterns, but their model interpretability is relatively poor. Additionally, they consume substantial computational resources, and the training process can be time-consuming.

\subsubsection{Health Rumor Detection Methods}
The mainstream methods still primarily utilize machine learning techniques, focusing on the unique characteristics of health rumors compared to general rumors. These characteristics include linguistic statistical features~\cite{20}, network features~\cite{21}, user features~\cite{22,23}, interaction features~\cite{23}, and sentiment features~\cite{25}, each possessing distinct traits. Small datasets collected autonomously are used for classification testing. Meanwhile, deep learning-based methods are also gradually gaining attention from researchers in the field of health rumor detection. For instance, Yang J. et al. propose and verify the rumor detection method based on content and user responses within a limited time, named CRLSTM-BE~\cite{26}. Rani P. et al. use the Bi-directional Long Short Term Memory (Bi-LSTM) model to prevent the spread of new rumors by continuously monitoring incoming messages in the network~\cite{27}.

\subsection{Large Language Models} 

Since the release of ChatGPT by OpenAI~\cite{Few-Shot_20_NIPS}, large language models (LLMs) have gained significant attention for their capabilities in natural language processing. Currently, LLMs are typically defined as Transformer-based neural network language models containing billions or more parameters and trained on massive datasets~\cite{LLM_survey_23_arXiv, LLM_survey_24_arXiv}, such as GPT-4~\cite{gpt4_24_arXiv}, LLaMA~\cite{llama_23_arXiv}, and Qwen~\cite{Qwen_23_arXiv}. Compared to earlier neural language models (NLMs) and pre-trained language models (PLMs), LLMs follow scaling laws~\cite{scaling_law_20_arXiv,training_22_arXiv}, and certain capabilities significantly improve when the model's parameter size exceeds a certain threshold, known as emergent abilities~\cite{emergent_22_arXiv}. The main emergent abilities of LLMs include (1) in-context learning (ICL)~\cite{ICL_survey_23_arXiv, Few-Shot_20_NIPS}, where LLMs can learn new tasks from a few examples provided in the prompt without needing retraining, (2) instruction following, where LLMs can understand and execute tasks based on natural language instructions, (3) multi-step reasoning, where LLMs can perform complex reasoning tasks by breaking them down into steps using methods like chain-of-thought (CoT)~\cite{CoT_22_NIPS}. Additionally, LLMs can be enhanced through techniques such as Retrieval-Augmented Generation (RAG)~\cite{RAG_survey_23_arXiv}, supervised fine-tuning (SFT)~\cite{finetuned_22_ICLR}, reinforcement learning from human feedback (RLHF)~\cite{RLHF_17_NIPS}, making them more robust and reliable for handling complex tasks in specialized fields. Consequently, numerous specialized LLMs have emerged in various vertical domains, such as Med-PaLM in the medical field~\cite{Med-PaLM_23_Nature}, BloombergGPT in finance~\cite{BloombergGPT_23_arXiv}, and ChatLaw in legal applications~\cite{chatlaw_23_arXiv}.

\subsection{Retrieval Augmented Generation}

Although LLMs currently demonstrate powerful language understanding and generation capabilities, they still struggle to adequately address questions requiring specialized domain knowledge or long-tail knowledge~\cite{long-tail-knowledge_23_ICML}, and often exhibit hallucination issues~\cite{Hallucination_23_arXiv}. Retrieval Augmented Generation (RAG) can effectively enhance LLMs and mitigate these problems~\cite{C-RAG_24_arXiv}. The RAG process mainly consists of three steps: indexing, retrieval, and generation. This involves collecting and processing external knowledge documents to construct a database, then using retrieval algorithms to recall documents relevant to the query from the database, and finally inputting the documents along with the original query into the LLM to generate a response~\cite{RAG_survey_23_arXiv}. Currently, researchers have proposed methods such as Query Optimization~\cite{HyDE_23_ACL, Query-Rewriting_23_EMNLP, LTM_23_arXiv} and Re-ranking~\cite{Reranking_23_EMNLP, Re2G_22_ACL} to optimize the retrieval of relevant documents, thereby further enhancing the performance of RAG.

The evaluation of RAG is divided into two main aspects: retrieval-based and generation-based~\cite{RAG_survey_24_arXiv}. The retrieval-based aspect focuses on the relevance and supportiveness of retrieved documents, using metrics such as Hit Rate, Context Relevance~\cite{ragas_24_ACL}, Error Detection Rate~\cite{benchmarking_rag_23_arXiv}. The generation-based aspect focuses on the quality of the generated answers, with common metrics including BLEU~\cite{bleu_02_ACL}, ROUGE~\cite{rouge_04_ACL}, EM, Micro-F1, Answer Relevance~\cite{ragas_24_ACL}.

\section{HRDE}
\label{sec:HRDE}
\subsection{Framework of HRDE}

\begin{figure}[ht]
  \centering
  \includegraphics[width=\linewidth]{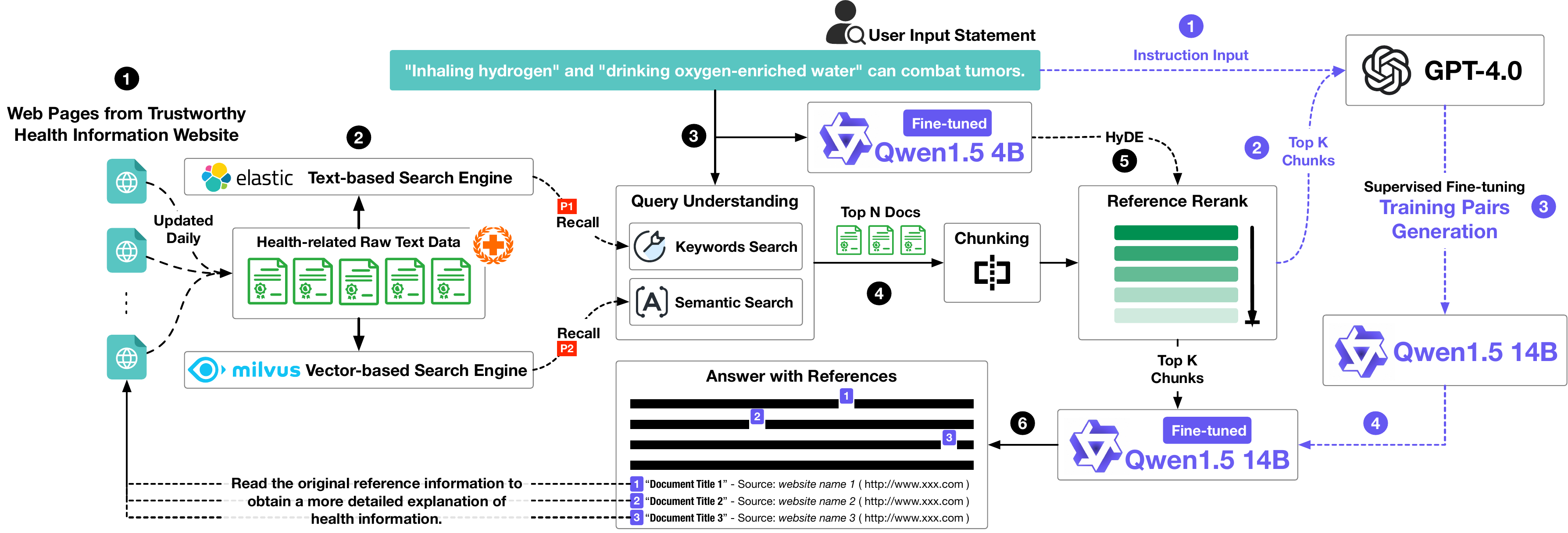}
  \caption{Framework of HRDE. The sections labeled with black numbers correspond to the process of health rumor detection, while the sections labeled with purple numbers correspond to the process of fine-tuning the large model.}
  \Description{An overall framework diagram of the HRDE method.}
  \label{fig:framework of HRDE}
\end{figure}

The proposed retrieval-augmented large language models for Chinese health rumor detection and explainability (HRDE) mainly consists of four parts: collection and storage of reference documents (\circled{black}{white}{1}\circled{black}{white}{2}), the retrieval of reference document (\circled{black}{white}{3}\circled{black}{white}{4}), the re-ranking of reference document (\circled{black}{white}{5}), and rumor detection answer generation by LLMs (\circled{black}{white}{6}). These components correspond to the processes labeled with black numbers in Figure ~\ref{fig:framework of HRDE}.

Firstly, the framework retrieves health-related information from multiple trustworthy websites and stores it in two reference databases (Elasticsearch and Milvus), which are updated daily (see Section ~\ref{sec:Data for Retrieval-Augmented Generation} for details). Then, based on the keywords and embedding vector of the user's input content, relevant reference documents are recalled from Elasticsearch and Milvus, respectively. Subsequently, Qwen1.5-4B-Chat is used to generate a hypothetical document to re-rank the recalled document chunks, and the top-k document chunks are selected as the final reference content (see Section ~\ref{sec:Evidence Recall} for details). Finally, the user input and the reference document chunks are input into Qwen1.5-14B-Chat to generate a response, which undergoes necessary post-processing.

The sections labeled with purple numbers in Figure ~\ref{fig:framework of HRDE} correspond to the model fine-tuning process, which includes collecting and integrating input information (\circled{purple}{white}{1}) and reference documents (\circled{purple}{white}{2}), generating a fine-tuning dataset based on GPT-4 (\circled{purple}{white}{3}), and performing model fine-tuning (\circled{purple}{white}{4}). Both Qwen1.5-4B-Chat and Qwen1.5-14B-Chat will be fine-tuned through these steps.

In the subsequent sections, we will introduce the construction of the HealthRCN dataset in Section ~\ref{sec:Data Preparation}, which will serve as the source for the subsequent fine-tuning and evaluation datasets. Section ~\ref{sec:Supervised Fine-tuning} will describe the fine-tuning process of Qwen1.5-4B-Chat and Qwen1.5-14B-Chat, with the fine-tuned models becoming core components of HRDE. Section \ref{sec:Retrieval-Augmented Generation} will detail the construction process of HRDE.

\subsection{Data Preparation}
\label{sec:Data Preparation}

\begin{figure}[ht]
  \centering
  \includegraphics[width=\linewidth]{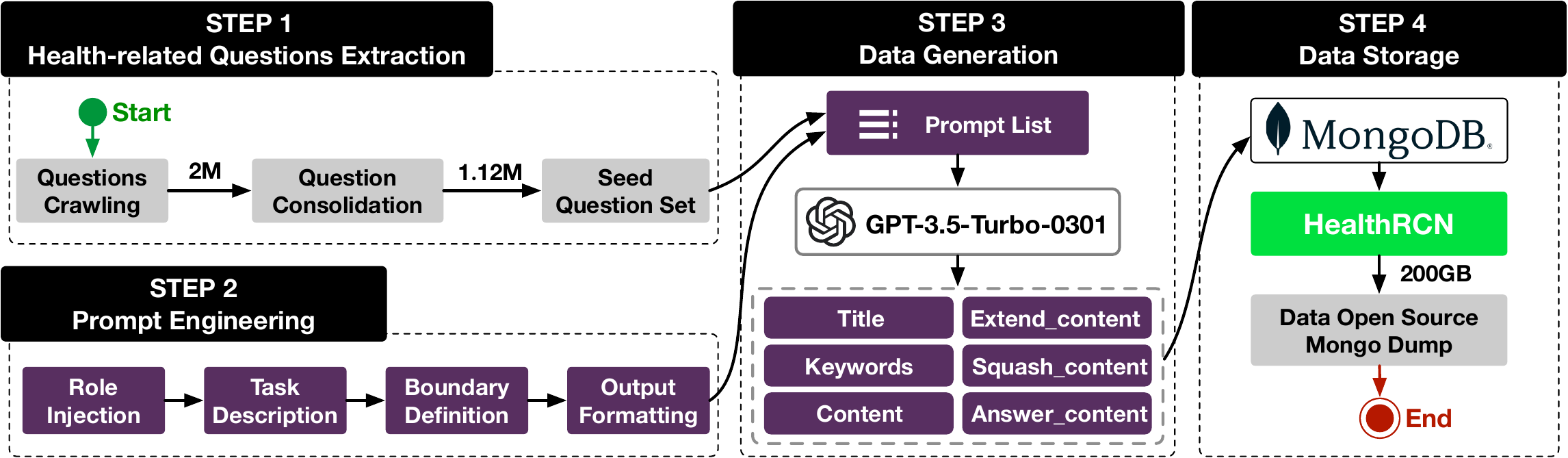}
  \caption{The Construction Process of HealthRCN}
  \Description{A detailed construction flowchart of HealthRCN.}
  \label{fig:HealthRCN}
\end{figure}

To address the gap in Chinese rumor datasets within the health domain, this paper constructs a dataset containing 1.12 million Chinese health-related rumors, named Health Rumor CN (HealthRCN, as shown in Figure \ref{fig:HealthRCN}). This dataset is the largest Chinese health information rumor dataset known to date.

To construct the HealthRCN dataset, we first gathered approximately 2 million user inquiries by crawling common questions from health-related websites\footnote{We primarily obtain health-related questions posted by users from \url{https://www.39.net/}.}. After consolidating these inquiries, we obtained 1.12 million health questions. Based on these health questions, we designed appropriate prompts (including role definitions, task descriptions, boundary definitions, and output formats) and used the GPT-3.5-Turbo-0301 model to sequentially generate corresponding rumor titles, keywords, short rumor content, long rumor content, refutation responses, and correct answers for each health question, consuming a total of approximately 3.36 billion tokens. The detailed information of the HealthRCN dataset fields is shown in Table ~\ref{tab:HealthRCN}. In Appendix ~\ref{sec:Prompt Used for Constructing the HealthRCN}, we show the prompts used to generate the corresponding fields, while in Appendix ~\ref{sec:Example of the HealthRCN}, we present a sample from this dataset.

Among these, the rumor titles, short rumor content, and long rumor content are health rumor titles and content generated by the GPT-3.5 based on the original health questions, with the long rumor content being an extended version of the short rumor content. The keywords field consists of all keywords extracted from the original health questions, separated by commas. The refutation responses clarify and explain the generated rumor content, while the correct answers address the original health questions accurately.

\begin{table}[ht]
\caption{Fields of HealthRCN}
\label{tab:HealthRCN}
\begin{minipage}{\columnwidth}
\begin{center}
\begin{tabular}{@{}llc@{}}
\toprule
Fields             & Description                                          & \multicolumn{1}{l}{Average length} \\ 
\midrule
Original\_question & Original health question                             & 11.1                               \\
Title              & The title of the generated rumor                     & 18.3                               \\
Keywords           & Keywords list                                         & 22.2/5.1                           \\
Content            & Generated short rumor content                        & 194.2                              \\
Extend\_content    & Generated extended rumor content                     & 579.1                              \\
Squash\_content    & Refutation response to generated rumor               & 366.5                              \\
Answer\_content    & The correct response to the original health question & 606.2                              \\ 
\bottomrule
\end{tabular}
\end{center}
\bigskip
    \footnotesize\emph{Note:} The "Average length" field indicates the average character length of the corresponding field text content, with the "Keywords" field specifying both the average length of keyword texts and the average number of keywords.
\end{minipage}
\end{table}

\subsection{Supervised Fine-tuning}
\label{sec:Supervised Fine-tuning}

To enable Qwen1.5-4B-Chat and Qwen1.5-14B-Chat to detect and respond to health-related rumors from user input in a coherent format, it is necessary to fine-tune both models in a supervised manner. This section will provide a detailed introduction to the datasets used for fine-tuning and the specifics of the fine-tuning process.

\subsubsection{Data for Fine-tuning}
\label{sec:Data for Fine-tuning}

The SFT dataset we used primarily consists of two categories of instruction data: question-answer data aimed at rumor detection and analysis, and conventional medical knowledge question-answer data. The former is used to guide the LLMs on how to perform the task of identifying health rumor and respond in a predetermined format. The latter serves to enhance the model's inherent medical knowledge and, on the other hand, to prevent overfitting of the question-answer format for rumor detection during the fine-tuning process.

The question-answer data for rumor detection and analysis is sourced from the HealthRCN dataset, which includes 21,246 question-answer pairs with reference documents and 10,000 question-answer pairs without reference documents. We randomly selected 31,246 health-related questions (original\_question) from the HealthRCN dataset and used GPT-4-1106-Preview to generate corresponding answers for rumor detection and analysis. The prompt template used to query GPT-4-1106-Preview is essentially the same as the one used for generating answers in the HRDE framework (see Section~\ref{sec:Interpretation Generation} for details). However, to ensure the responses generated by GPT-4-1106-Preview meet our expectations, we included an additional question-answer example (1-shot) in the prompt. The reference document retrieval strategy is also consistent with the method used in the HRDE framework. Due to the absence of a fine-tuned Qwen1.5-4B-Chat model, the re-ranking process was simplified. The complete method is detailed in Section~\ref{sec:Evidence Recall}.

The conventional medical knowledge question-answering data comes from the following five open-source Chinese medical instruction datasets: ChatMed Consult Dataset~\cite{ChatMed-Dataset_23_github}, ShenNong TCM Dataset~\cite{ShenNong-TCM_23_github}, Chinese medical dialogue data\footnote{\url{https://github.com/Toyhom/Chinese-medical-dialogue-data}}, Huatuo-Llama-Med-Chinese Dataset~\cite{HuaTuo_23_arxiv}, and Med-ChatGLM Dataset~\cite{ChatGLM-Med_23_github}. These five open-source instruction sets consist of simple medical knowledge question-answer instructions. We randomly selected 6,000 data entries from each dataset, resulting in a total of 30,000 conventional medical knowledge question-answer data entries.

Finally, the aforementioned two types of data are processed into a unified format, then merged and shuffled to form the final SFT dataset (see Table ~\ref{tab:Fine-tuning Dataset} for details), which contains 61,246 question-answer pairs.

\begin{table}[ht]
\caption{Fine-tuning Dataset}
\label{tab:Fine-tuning Dataset}
\begin{minipage}{\columnwidth}
\begin{center}
\begin{tabular}{@{}lc@{}}
\toprule
Data type                          & Sample size  \\ 
\midrule
Refutation Q\&A with references    & 21246 \\
Refutation Q\&A without references & 10000 \\
ChatMed Consult Dataset            & 6000  \\
ShenNong TCM Dataset               & 6000  \\
Chinese medical dialogue data      & 6000  \\
Huatuo-Llama-Med-Chinese Dataset   & 6000  \\
Med-ChatGLM Dataset                & 6000  \\ 
\bottomrule
\end{tabular}
\end{center}
\end{minipage}
\end{table}

\subsubsection{Model Fine-tuning}

To enable the model to better detect and analyze health-related rumors from user inputs and output responses in a predefined format, we fine-tuned Qwen1.5-4B-Chat and Qwen1.5-14B-Chat models using the dataset constructed in Section~\ref{sec:Data for Fine-tuning}. 

For the Qwen1.5-14B-Chat model, we employed the LoRA method~\cite{LoRA_22_ICLR} for efficient fine-tuning, setting the rank of LoRA to 16. Specifically, we fine-tuned the Qwen1.5-14B-Chat model on seven NVIDIA A800-SXM4-40GB GPUs for 3 epochs, which took approximately 4.5 hours (16264.3 seconds). 

For the Qwen1.5-4B-Chat model, we adopted a full fine-tuning approach. Specifically, we fine-tuned the Qwen1.5-4B-Chat model on two NVIDIA H800 PCIe GPUs for 2 epochs, which took approximately 4.7 hours (16904.7 seconds). 


\subsection{Retrieval-Augmented Generation}
\label{sec:Retrieval-Augmented Generation}

\subsubsection{Data for Retrieval-Augmented Generation}
\label{sec:Data for Retrieval-Augmented Generation}

To build the reference database required for RAG, we primarily used Elasticsearch and Milvus to store and retrieve health-related reference documents. We sourced information to construct the reference database from 19 trustworthy websites or professional organizations (as detailed in the Appendix ~\ref{sec:Trustworthy health information website}). To ensure the timeliness and accuracy of the reference database, we regularly update it by fetching new data daily from websites that maintain updated information.

For each health information article from these sources, we capture the title, body content, publication date, source name, and URL. This data is then duplicated and stored in both Elasticsearch and Milvus databases, providing comprehensive information and traceability for subsequent use. In the Elasticsearch, we store the complete article as a single record (i.e., a document in Elasticsearch). However, for Milvus, due to the context window length limitation of the m3e embedding model we use (512 tokens), we segment the body of each article into chunks (ensuring each chunk is less than 512 tokens in length) and generate corresponding embedding vectors using the embedding model. Each chunk, along with its embedding vector, article title, publication date, and URL, is then sequentially stored in Milvus.

As of the completion of this paper, Elasticsearch stores information on 358,114 reference documents, while Milvus stores information on 1,347,976 reference document chunks.

\subsubsection{Evidence Recall}
\label{sec:Evidence Recall}

The recall of reference documents in the HRDE method primarily consists of two parts: the recall and re-ranking of reference documents.

First, we recall reference documents related to the user input from both the Elasticsearch and Milvus databases simultaneously. For Elasticsearch, we extract keywords from the user input to recall reference documents. As a traditional database, Elasticsearch uses inverted indexing and the BM25 algorithm, which effectively supports keyword-based document retrieval. For the vector database Milvus, we use the embedding vector of the user input (generated by the m3e embedding model) to recall reference document chunks. Using both keywords and embedding vectors for recall ensures both the accuracy and coverage of the recalled documents, thereby fully leveraging the effective information in the reference database to assist the LLM in generating answers. Since the reference documents recalled from Elasticsearch are complete articles, they need to be chunked. Then, the reference document chunks recalled from both sources are merged to form the Top-N chunks.

Next, for the Top-N document chunks recalled from both sources, we use HyDE (Hypothetical Document Embeddings)~\cite{HyDE_23_ACL} for re-ranking. Re-ranking is necessary for two main reasons: LLM's context window length is limited, preventing it from processing too many reference document chunks at once, and there is a need to further filter the reference document chunks to remove less relevant or noisy chunks.

Specifically, we used the fine-tuned Qwen1.5-4B-Chat to independently generate a response to the user input's health information query without using reference documents as a hypothetical document. We then calculate the semantic similarity between all the recalled document chunks and the hypothetical document (evaluated using cosine similarity) and rank them, selecting the Top-K document chunks as the final reference content. Re-ranking based on HyDE helps eliminate some document chunks that are related to the user input but do not contribute to rumor analysis, thereby ensuring the quality of the final reference document chunks. Additionally, we set a semantic similarity threshold in the re-ranking process to directly exclude document chunks with very low similarity to the hypothetical document.

\subsubsection{Interpretation Generation} \label{sec:Interpretation Generation}

To enable the LLM to better identify rumors based on user input, we designed two types of prompts to guide the model in providing rumor detection responses, both with and without reference documents. The non-referenced prompt is also used by Qwen1.5-4B-Chat to generate the hypothetical document. Each type of prompt consists of four parts: task description, input information, detailed task requirements, and answer output format.

\begin{figure}[ht]
  \centering
  \includegraphics[width=\linewidth]{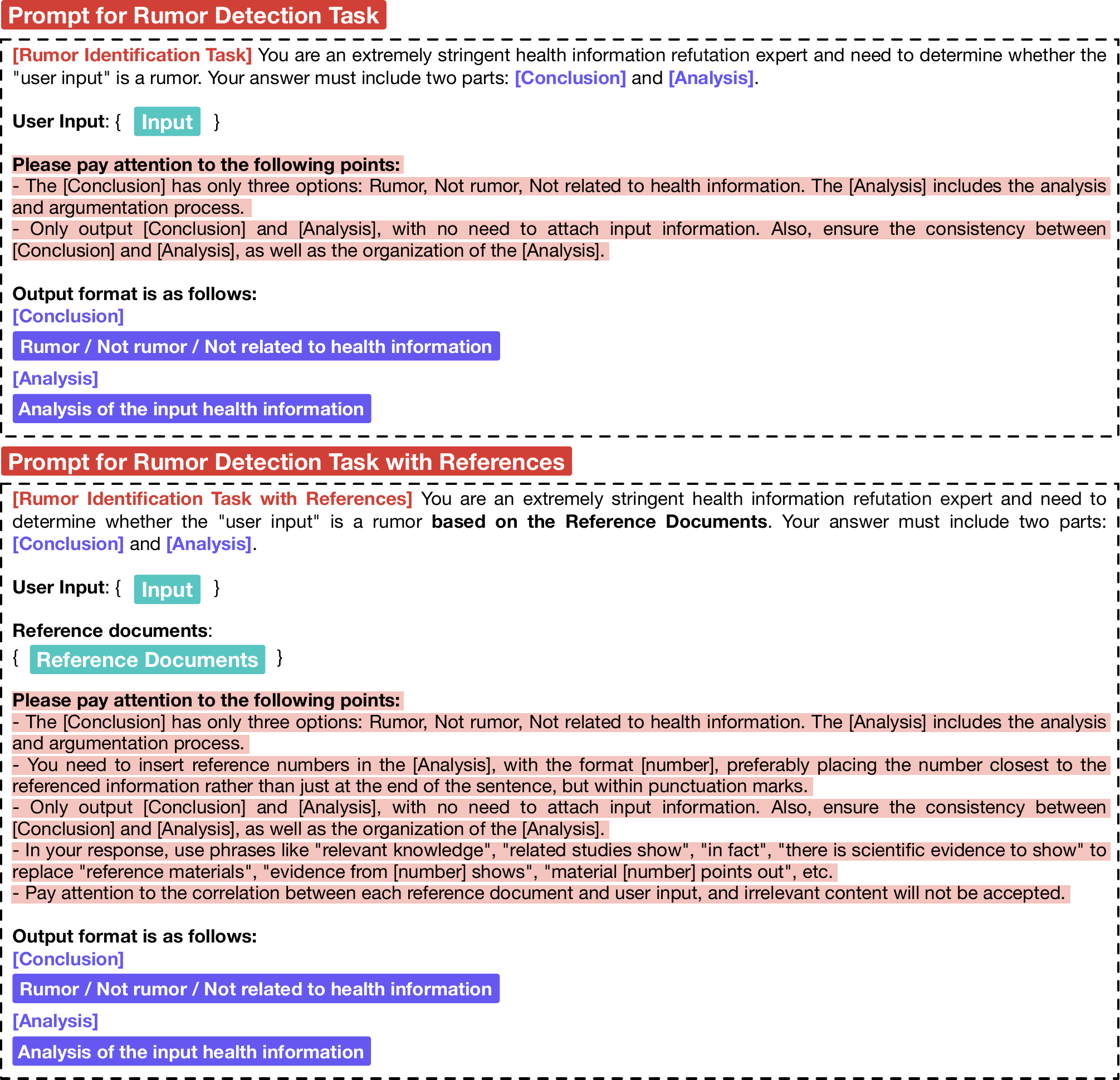}
  \caption{Prompts for Refutation}
  \Description{Prompts for Refutation.}
  \label{fig:Prompts for Refutation}
\end{figure}

The task description briefly outlines the content of the rumor detection task. We use the identifiers [Rumor Detection Task] and [Rumor Detection Task with References] to indicate whether the current task will include reference documents. In the input information section, the user's health information input is embedded. If the current task requires reference documents, the recalled reference documents are also embedded sequentially in this section. The final two parts provide detailed task requirements and the specific format for the response, ensuring standardized answers from the model (see Figure ~\ref{fig:Prompts for Refutation} for the English translations of the prompts, with the original Chinese versions available in the Appendix ~\ref{sec:Original Prompts for Refutation}).

After obtaining the reference documents, we will embed the user's input and the reference documents (if needed) into a designed prompt, which will then be processed by the fine-tuned Qwen1.5-14B-Chat to provide a response for rumor detection. Additionally, if no valid documents are retrieved or all retrieved documents are excluded during the re-ranking phase, it indicates that there is no available information in the database. In this case, we will use the prompt without reference documents to guide the LLM, allowing it to respond directly based on its internal knowledge.

The response for rumor detection will include three sections: conclusion, analysis, and references. According to the requirements in the prompt, the labels in the conclusion can only include "Rumor," "Not rumor," and "Not related to health information." The analysis section will analyze the input health information, providing a complete reasoning and analysis process. If there are reference documents available, the model will use content from the reference documents for reasoning and validation in the analysis section, with citations indicated in the text (e.g., “[1]”). Finally, if the analysis section uses content from the reference documents, the source information for these references will be provided in the references section at the end of the response. The first two sections are generated directly by the model, while the third section will be supplemented during post-processing based on the model’s response and the reference documents. Specific Q\&A examples can be found in the Section ~\ref{sec: Case Study}.

\section{Experiment}
\label{sec:Experiment}
\subsection{Base Model Selection}
\label{sec:Base Model Selection}

The LLMs utilized in the experiment include ChatGLM3-6B~\cite{GLM_22_ACL_CCFA}, Baichuan2-13B-Chat~\cite{Baichuan2_23_arXiv_Baichuan}, Qwen1.5-14B-Chat~\cite{Qwen_23_arXiv}, GPT-3.5-Turbo\footnote{\url{https://openai.com}}, and GPT-4-1106-Preview\footnote{\url{https://openai.com/blog/new-models-and-developer-products-announced-at-devday}}. These models encompass both open-source and proprietary models that are currently representative in the field.

The experiments in this paper primarily designed the following four types of models for comparative analysis:

\textbf{LLM.} The model utilizes the original LLM for rumor detection and analysis tasks, serving as a baseline for comparison with other models.

\textbf{LLM + RAG.} The model extends the LLM by incorporating the RAG framework designed in this paper, enabling the model to reference recalled relevant documents for refuting rumors. 

\textbf{LLM + SFT.} The model utilizes a fine-tuned LLM. LLMs of similar parameter scale are fine-tuned using identical methods and parameters, as detailed in Section ~\ref{sec:Supervised Fine-tuning}. Through this fine-tuning process, the model learns rumor detection techniques and acquires some knowledge specific to the healthcare domain from the fine-tuning dataset. Additionally, its output becomes more standardized.

\textbf{LLM + SFT + RAG.} The model utilizes a fine-tuned LLM, integrated with the RAG framework designed in this paper. It will be compared with the three preceding models in ablation experiments, thereby validating the effectiveness and necessity of the fine-tuning and RAG components designed in this paper.

The HRDE model proposed in this paper is Qwen1.5-14B-Chat + SFT + RAG. For models requiring RAG, the hypothetical documents in the RAG process are generated by the fine-tuned Qwen1.5-4B-Chat. Additionally, we configure the retrieval of 5 complete reference documents from Elasticsearch, and 25 reference document chunks from Milvus. During the re-ranking phase, we select the final five reference document chunks, with a semantic similarity threshold (based on cosine similarity) set at 0.5.

\subsection{Datasets}

The evaluation dataset primarily comes from the HealthRCN dataset, supplemented with some evaluation samples unrelated to health information (constructed manually), totaling 2500 evaluation samples. The sample labels in the evaluation dataset are categorized into three types: "Rumor" (1497), "Not rumor" (855), and "Not related to health information" (148).For samples from the HealthRCN dataset, we employed a random sampling strategy. We ensured that the selected samples did not overlap with those used for fine-tuning, thereby maintaining the independence of the evaluation dataset. For rumor samples, we use the "Content" field as the input content, and for non-rumor samples, we use the "Answer\_content" field as the input content.

\subsection{Model Deployment and Inference}


This paper employs the vLLM framework for deploying LLMs. vLLM is an open-source framework for LLM deployment that supports fast inference and serving of LLMs~\cite{Vllm_23_SOSP}. vLLM utilizes a novel attention algorithm called PagedAttention, which efficiently manages the keys and values of attention. Additionally, this framework supports the deployment of numerous LLMs, including the Qwen1.5-14B-Chat and Qwen1.5-4B-Chat models used in this paper.

And the Qwen1.5-4B-Chat model and the Qwen1.5-14B-Chat model were deployed using the vLLM framework on the NVIDIA H800 PCIe. The inference speeds reached 698.5 tokens per second and 143.6 tokens per second, respectively, while the serving throughput, with each request asking for ten parallel output completions, achieved 153.8 requests per minute and 68.6 requests per minute, respectively.

\subsection{Evaluation Metrics}

The core evaluation metrics in this study include accuracy, F1 score, relevance, reliability, and richness. Accuracy and F1 score measure the accuracy of the model's predicted labels, assessing whether the model correctly classifies the input health information. The F1 score considers both precision and recall to provide a balanced evaluation. Relevance, reliability, and richness primarily assess the quality of the model's answer.

Relevance evaluates whether the model's answer focuses on the input health information and addresses whether the information is a rumor. Reliability assesses whether the model's analysis content is convincing, supporting the predicted label with scientific evidence and reasonable inferences, avoiding language that could cause new misunderstandings or confusion. Richness evaluates the diversity and coverage of the model's analysis content, determining if the analysis includes comprehensive information and arguments, and if it covers all key points or aspects mentioned in the input.

Additionally, this study records the proportion of valid responses from each model. A valid response follows the format given in the prompt and provides a predicted label within the specified range. The calculations for accuracy and F1 score only consider the evaluation samples with valid responses, excluding those with invalid responses. However, relevance, reliability, and richness evaluations include all model responses, regardless of their validity.

The objective of the rumor refutation task in this paper is two-fold: to determine whether the input information is a rumor and to provide an analysis for the judgment. Consequently, different evaluation metrics are required to measure the model's performance in achieving these two goals. Accuracy and F1 score are used to gauge the model's success in detecting rumors, while relevance, reliability, and richness are employed to assess the model's effectiveness in providing the analysis. Additionally, the proportion of valid responses can be used to evaluate the model's ability to comply with instructions.

Due to the difficulty of calculating relevance, reliability, and richness through explicit formulas, we used the GPT-4-1106-Preview model to evaluate the model's responses. We designed specific prompts (detailed in the Appendix ~\ref{sec:Prompt for Evaluation}) for GPT-4-1106-Preview to sequentially score each response on relevance, reliability, and richness. Each metric is scored on a scale of 0-10, with higher scores indicating better performance on the respective metric. Finally, we calculate the average score to serve as the model's final rating for each metric.

\subsection{Evaluation Results}

Table ~\ref{tab:Ablation Experiment} and Table ~\ref{tab:Multiple LLM Evaluation Experiments} present all evaluation results, including the ablation experiment results and the comparative evaluation results of multiple LLMs. Overall, our proposed HRDE model outperforms all other models in nearly all evaluation metrics, validating its effectiveness and superiority. In terms of rumor detection accuracy, HRDE achieves an average accuracy of 91.04\% and an F1 score of 91.58\%, significantly surpassing the other models. Additionally, HRDE exceeds GPT-4-1106-Preview by more than 2\% in both of these metrics. This indicates that HRDE can classify and identify rumors more accurately, maintaining a good balance between precision and recall. Regarding answer quality, the HRDE model consistently ranks either first or second across the relevance, reliability, and richness metrics. This demonstrates that HRDE's answer can better incorporate reference documents to analyze user input effectively. In terms of relevance, HRDE is second only to GPT-4-1106-Preview and Qwen1.5-14B-Chat + SFT. This is mainly because HRDE relies more on the reference documents provided by RAG, significantly enhancing the richness of the answers, which slightly lowers the relevance score. Furthermore, in terms of valid answer rate, HRDE can perfectly output answers in the format specified by the prompt, ensuring the consistency of each answer.

\begin{table*}[h]
    \caption{Ablation Experiment}
    \label{tab:Ablation Experiment}
    \resizebox{\linewidth}{!}{
    \begin{threeparttable}
    \begin{tabular}{@{}lcccccccc@{}}
    \toprule
    \multicolumn{1}{c}{\textbf{Model}} & \multicolumn{1}{c}{\textbf{SFT}} & \multicolumn{1}{c}{\textbf{RAG}} & \textbf{Avg Accuracy} & \textbf{F1 Score} & \textbf{Valid Answer Rate} & \textbf{Relevance} & \textbf{Reliability} & \textbf{Richness} \\ \midrule
    \textbf{Qwen1.5-14B-Chat} &  &  & 57.37\% & 46.93\% & 62.68\% & 8.77 & 8.16 & 6.86 \\
    \textbf{Qwen1.5-14B-Chat} & \checkmark &  & \underline{89.64\%} & \underline{90.80\%} & \textbf{100.00\%} & \textbf{9.27} & \underline{8.89} & \underline{6.93} \\
    \textbf{Qwen1.5-14B-Chat} &  & \checkmark & 56.79\% & 46.40\% & \underline{82.76\%} & 7.92 & 7.88 & 6.65 \\
    \textbf{HRDE} & \checkmark & \checkmark & \textbf{91.04\%} & \textbf{91.58\%} & \textbf{100.00\%} & \underline{9.25} & \textbf{8.99} & \textbf{7.3} \\ 

    \bottomrule
    \end{tabular}
    \begin{tablenotes}
            \footnotesize
            \item \textit{Note}: "HRDE" specifically refers to Qwen1.5-14B-Chat + SFT + RAG. The best performance in each column will be bolded, and the second-best performance will be underlined.
    \end{tablenotes}
    \end{threeparttable}
    }
\end{table*}

In the ablation experiment (Table ~\ref{tab:Ablation Experiment}), HRDE achieves the best performance on almost all metrics compared to the Qwen1.5-14B-Chat model that does not use SFT and RAG simultaneously. This indicates that using SFT or RAG alone is not the optimal strategy. When only RAG is used, the model, lacking fine-tuning, cannot effectively leverage information from reference documents to improve answer quality. In fact, it may be misled by the information, leading to a decline in most metrics compared to the model without RAG. On the other hand, when only SFT is employed, the model's rumor detection accuracy and valid answer rate significantly improve (though still not surpassing HRDE), and the relevance of answers even exceeds that of HRDE. However, the model exhibits deficiencies in reliability and richness, as the absence of external information from reference documents means it cannot provide more credible evidence and knowledge to support and enrich its answers. Therefore, combining SFT and RAG ensures significant improvements in both rumor detection and answer quality. Additionally, in terms of the valid answer rate, models using RAG show a significant enhancement compared to the original Qwen1.5-14B-Chat. This improvement is mainly due to the HyDE method used in RAG for filtering reference documents, where the hypothetical documents are generated by the fine-tuned Qwen1.5-4B-Chat model. This ensures that the re-ranked reference documents are primarily responses to the input query regarding the veracity of health information, thereby significantly improving the model's valid answer rate.

\begin{table*}[h]
    \caption{Multiple LLM Evaluation Experiments}
    \label{tab:Multiple LLM Evaluation Experiments}
    \resizebox{\linewidth}{!}{
    \begin{threeparttable}
    \begin{tabular}{@{}lcccccccc@{}}
    \toprule
    \multicolumn{1}{c}{\textbf{Model}} & \multicolumn{1}{c}{\textbf{SFT}} & \multicolumn{1}{c}{\textbf{RAG}} & \textbf{Avg Accuracy} & \textbf{F1 Score} & \textbf{Valid Answer Rate} & \textbf{Relevance} & \textbf{Reliability} & \textbf{Richness} \\ \midrule
    \textbf{GPT-3.5-Turbo} &  & \checkmark & 63.18\% & 53.68\% & 99.52\% & 7.06 & 6.53 & 3.18 \\
    \textbf{GPT-4-1106-Preview} &  & \checkmark & 88.20\% & \underline{88.15\%} & \textbf{100.00\%} & \textbf{9.27} & \underline{8.97} & \underline{6.71} \\
    \textbf{ChatGLM3-6B} & \checkmark & \checkmark & 88.01\% & 85.24\% & \underline{99.76\%} & 9.03 & 8.59 & 6.16 \\
    \textbf{Baichuan2-13B-Chat} & \checkmark & \checkmark & \underline{88.47}\% & 85.22\% & 92.64\% & 8.36 & 8.29 & 5.61 \\
    \textbf{HRDE} & \checkmark & \checkmark & \textbf{91.04\%} & \textbf{91.58\%} & \textbf{100.00\%} & \underline{9.25} & \textbf{8.99} & \textbf{7.3} \\ 

    \bottomrule
    \end{tabular}
    \begin{tablenotes}
            \footnotesize
            \item \textit{Note}: "HRDE" specifically refers to Qwen1.5-14B-Chat + SFT + RAG. The best performance in each column will be bolded, and the second-best performance will be underlined.
    \end{tablenotes}
    \end{threeparttable}
    }
\end{table*}

Table ~\ref{tab:Multiple LLM Evaluation Experiments} presents the comparative evaluation results among several different LLMs. Compared to ChatGLM3-6B and Baichuan2-13B-Chat models, which also employ both SFT and RAG, HRDE demonstrates significant advantages across all evaluation metrics. This is primarily due to the superior baseline performance of Qwen1.5-14B-Chat, which excels in natural language processing capabilities relative to ChatGLM3-6B and Baichuan2-13B-Chat. Consequently, Qwen1.5-14B-Chat, after fine-tuning, is better equipped to handle and utilize information from reference documents for health information detection and analysis.

Furthermore, HRDE significantly outperforms GPT-3.5-Turbo across all evaluation metrics, despite GPT-3.5-Turbo having a much larger parameter size than Qwen1.5-14B-Chat. Compared to the more advanced GPT-4-1106-Preview, HRDE still excels in almost all evaluation metrics, with the exception of being slightly lower in relevance of answers than GPT-4-1106-Preview. This demonstrates the effectiveness of the SFT and RAG designs proposed in this paper, which can significantly enhance the overall performance of Qwen1.5-14B-Chat in handling health rumor tasks.

\begin{table*}[h]
    \caption{Effect of Semantic Similarity Threshold on HRDE}
    \label{tab:Effect of Semantic Similarity Threshold on HRDE}
    \resizebox{\linewidth}{!}{
    \begin{threeparttable}
    \begin{tabular}{@{}cccccccc@{}}
    \toprule
    \textbf{Model} & \textbf{\makecell[c]{Similarity\\ Threshold}} & \textbf{Avg Accuracy} & \textbf{F1 Score} & \textbf{Relevance} & \textbf{Reliability} & \textbf{Richness} \\ 
    \midrule
    HRDE & 0.1 & \underline{91.44\%} & 90.39\% & \textbf{9.2728} & \textbf{9.0118} & \textbf{7.7130} \\
    HRDE & 0.3 & \textbf{91.52\%} & 90.88\% & \textbf{9.2728} & \textbf{9.0118} & \underline{7.6996} \\
    HRDE & 0.5 & 91.04\% & \textbf{91.58\%} & 9.2544 & \underline{8.9912} & 7.3399 \\
    HRDE & 0.7 & 89.64\% & 90.50\% & \underline{9.2733} & 8.8892 & 6.9323 \\
    HRDE & 0.9 & 89.76\% & \underline{90.99\%} & 9.2725 & 8.8884 & 6.9295 \\ 
    \bottomrule
    \end{tabular}
    \begin{tablenotes}
            \footnotesize
            \item \textit{Note}: "HRDE" specifically refers to Qwen1.5-14B-Chat + SFT + RAG. The best performance in each column will be bolded, and the second-best performance will be underlined.
    \end{tablenotes}
    \end{threeparttable}
    }
\end{table*}

In HRDE, an important parameter is the semantic similarity threshold, which filters out reference documents with low semantic similarity to the input information (see Section ~\ref{sec:Evidence Recall}), thereby adjusting the quality of the reference documents provided to the LLM. In previous experiments, this parameter was set to 0.5 (see Section ~\ref{sec:Base Model Selection}), meaning that reference documents with a semantic similarity less than 0.5 to the input information were excluded. In Table ~\ref{tab:Effect of Semantic Similarity Threshold on HRDE}, we present the performance of HRDE under different semantic similarity thresholds. As the semantic similarity threshold decreases, the model's average accuracy shows an upward trend, but the F1 score exhibits a downward trend. This indicates that a lower semantic similarity threshold can provide more relevant reference documents, improving the model's average accuracy. Still, the recall rate decreases, leading to a lower F1 score. Similarly, under stricter semantic similarity thresholds, the reference documents obtained by the model contain less noise, improving the quality of the answers, albeit to a lesser extent. Considering all evaluation metrics, a semantic similarity threshold of 0.5 will be an appropriate default value.

\section{Case Study}
\label{sec: Case Study}

This section will select three real-world Q\&A examples to demonstrate HRDE's ability to detect health rumor and provide corresponding interpretability analysis. The input information for these three cases consists of widely spread or highly impactful Chinese health rumors or medical common knowledge. In practice, both the health information and the original responses in the three examples are in Chinese. We have translated them into English for the convenience of the readers, and the original Chinese text will be presented in the Appendix ~\ref{sec:Original Q&A Text of the Case Study}.

\subsubsection{Case One: mobile phone radiation can cause cancer}

\begin{figure}[ht]
  \centering
  \includegraphics[width=\linewidth]{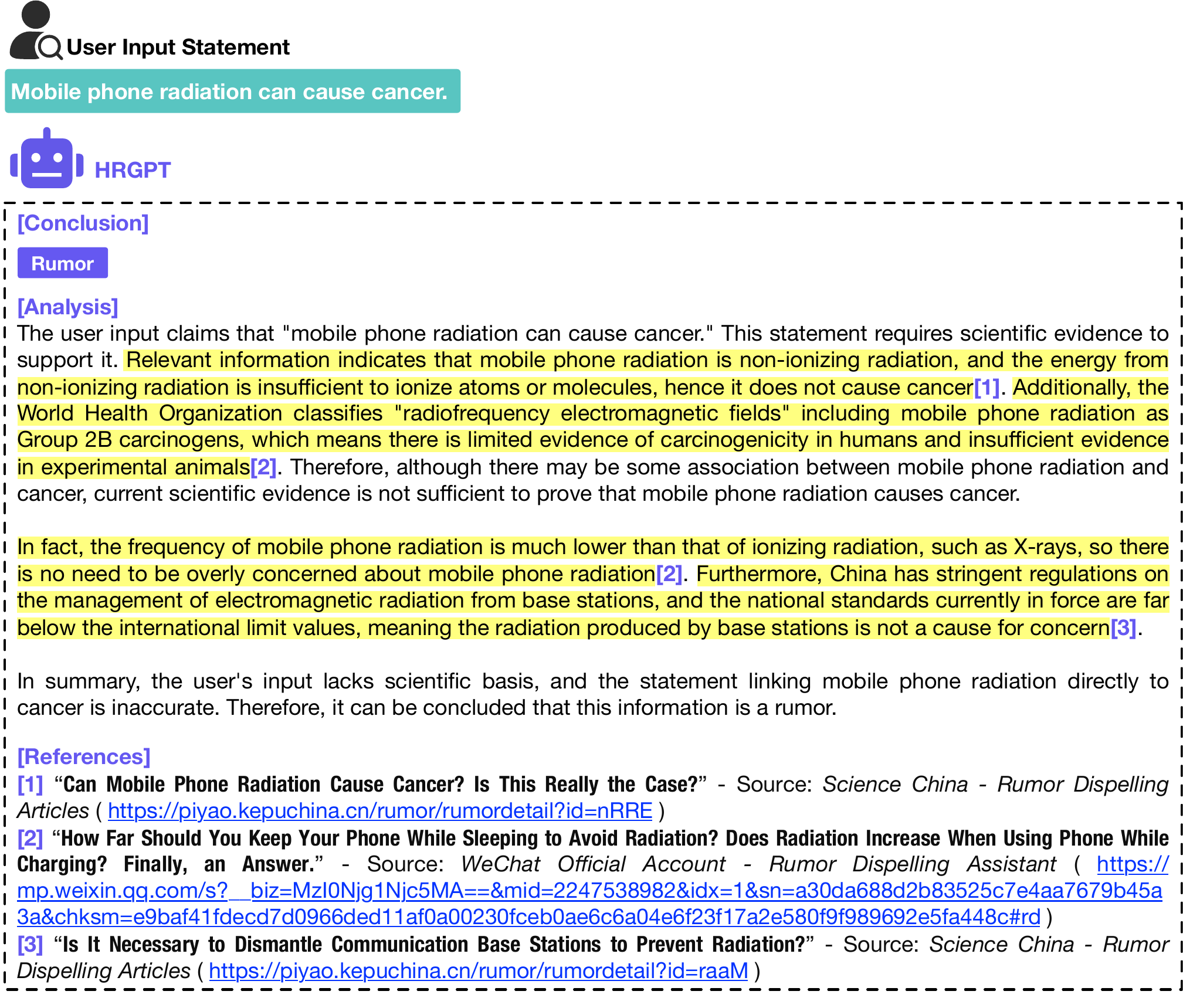}
  \caption{Case one: Mobile phone radiation can cause cancer.}
  \Description{Case: Mobile phone radiation can cause cancer.}
  \label{fig:Case One}
\end{figure}

The statement "mobile phone radiation can cause cancer" is a widely circulated health rumor. After inputting this statement into the HRDE model, the model's response, after post-processing, is shown in Figure ~\ref{fig:Case One}. The final response is divided into three main parts: the conclusion of the rumor detection, the analysis of the user's input information, and the source information of the reference documents.

The HRDE model explicitly states in the conclusion that it is a rumor, which is a correct judgment. In the analysis section, the model first emphasizes that such health information requires scientific evidence for support and should not be blindly trusted. It then cites content from the first and second reference documents to explain the principles of cell phone radiation and the viewpoint of professional organizations (World Health Organization) regarding cell phone radiation, arguing that there is currently insufficient scientific evidence to prove that cell phone radiation necessarily causes cancer. In the second paragraph, the model again references literature to indicate that the harm from cell phone radiation is much less than that from ionizing radiation, suggesting there is no need for excessive panic. Additionally, it explains that China has strict regulations for the base stations that transmit cell phone signals, alleviating concerns about the radiation they produce. In the third paragraph, the model provides a summary of the entire analysis, concluding that the health information is not credible and is a rumor. At the end of the response, it lists the sources of all reference documents, including three related articles from two websites.

In the model's response, we can observe that the analysis section mentions base stations related to mobile phone radiation, which were not referenced in the user's input (sourced from the third reference document). This type of information enhances the richness of the model's response and provides the user with additional relevant knowledge, helping them further understand related health information. At the same time, this is the reason why the HRDE model scores slightly lower in relevance compared to the LLM + SFT model. However, this does not negatively impact the user’s reading experience; instead, the inclusion of extensive relevant knowledge increases the overall reliability of the response.

\subsubsection{Case Two: vitamin C can prevent colds}

\begin{figure}[ht]
  \centering
  \includegraphics[width=\linewidth]{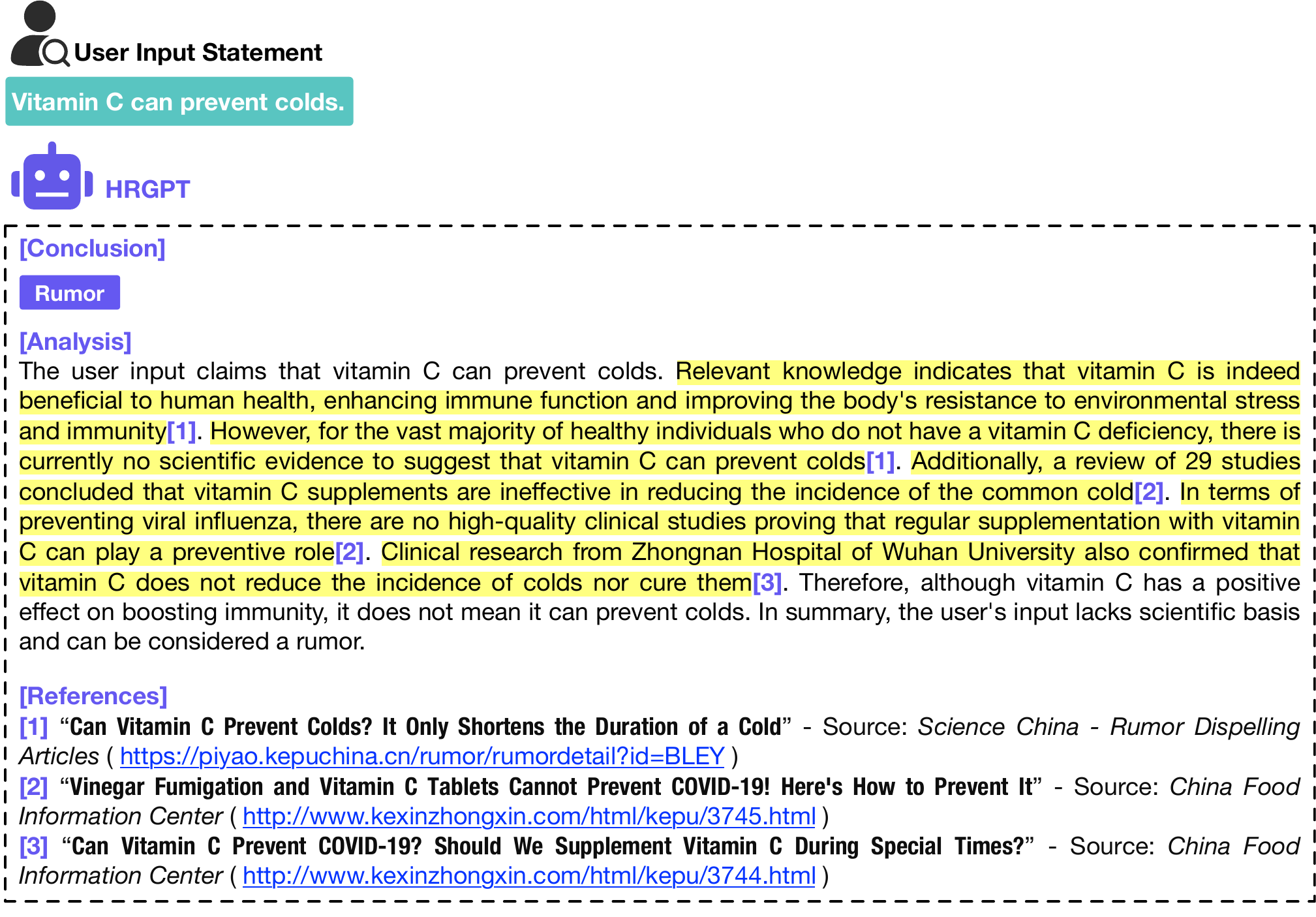}
  \caption{Case Two: Vitamin C can prevent colds.}
  \Description{Case: Vitamin C can prevent colds.}
  \label{fig:Case Two}
\end{figure}

"Vitamin C can prevent colds" is also an extremely widespread health myth. Figure ~\ref{fig:Case Two} shows the response of the HRDE model to this health information.

In the conclusion, the HRDE model gives the correct conclusion that this health information is a rumor. In the analysis section, the model first cites the first reference document to provide the general effects and benefits of vitamin C, but also explicitly states that for healthy individuals, vitamin C does not have the ability to prevent colds. It then cites the second reference document which explains that currently there is no professional evidence proving that vitamin C can effectively prevent common colds or viral influenza. Next, it cites the third reference document which provides the research conclusion of a professional medical institution (Wuhan University Zhongnan Hospital), stating that vitamin C cannot effectively prevent or treat colds. Finally, the model summarizes that while vitamin C can indeed boost the immune system, there is no clear evidence indicating that it can prevent colds. In the references section, the model's response provides the source information of the three reference documents used.

From the model's response, we can see that the model effectively utilizes and organizes information from all reference documents, explaining to the user in a clear and convincing manner, from the original efficacy of vitamin C to the research conclusions given by authoritative institutions, that vitamin C cannot prevent colds.

\subsubsection{Case Three: the main cause of diabetes is not eating too much sugar}

\begin{figure}[ht]
  \centering
  \includegraphics[width=\linewidth]{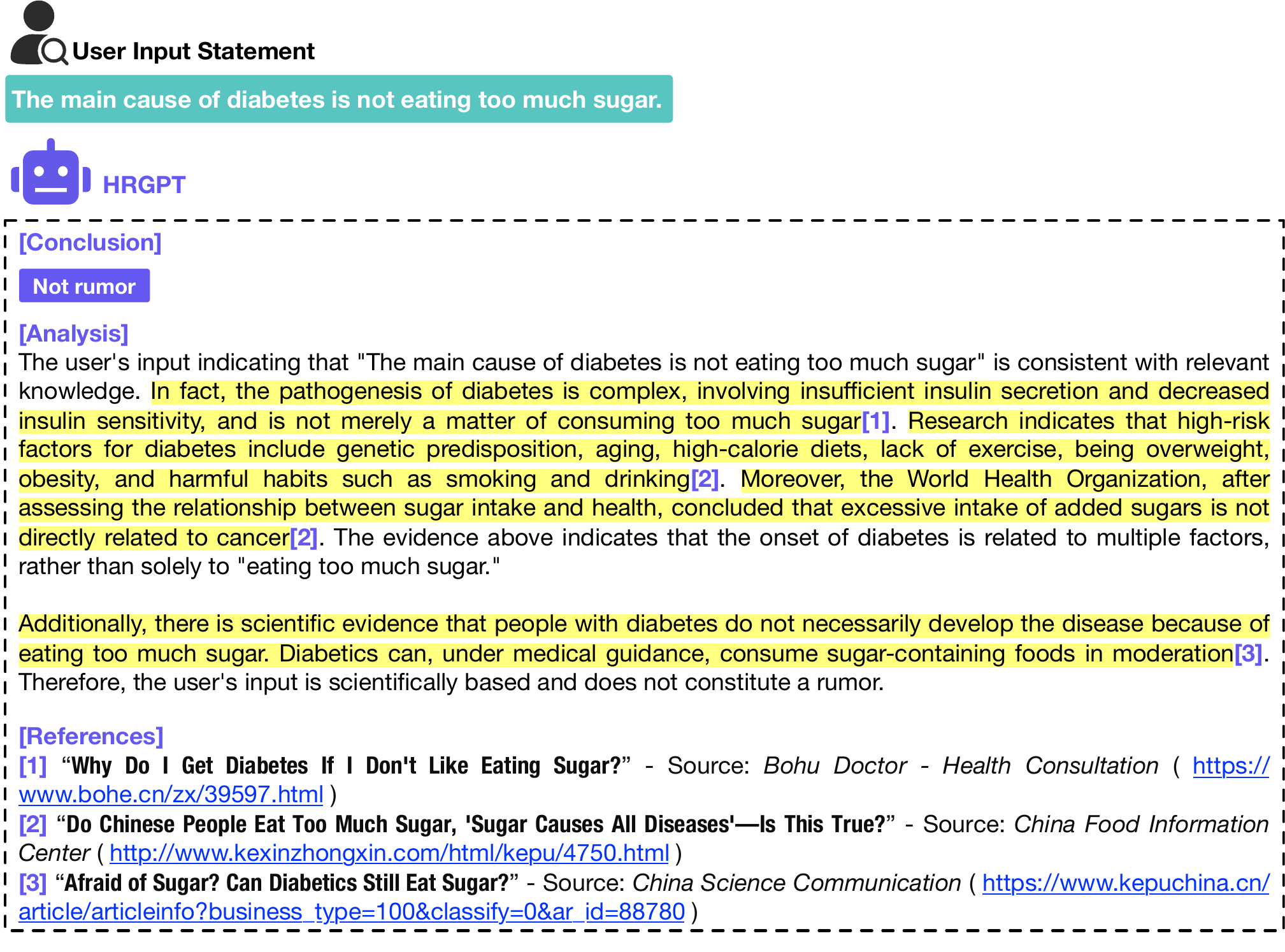}
  \caption{Case Three: The main cause of diabetes is not eating too much sugar.}
  \Description{Case: The main cause of diabetes is not eating too much sugar.}
  \label{fig:Case Three}
\end{figure}

"The main cause of diabetes is not eating too much sugar" is a widely disseminated and controversial health information. However, this statement is correct medical knowledge. We hope to explore the pattern of responses from the HRDE model when faced with accurate health information through this case.

In the conclusion section, the HRDE model gives the correct judgment that the input health information is not a rumor. In the analysis section, the model first indicates that the pathogenesis of diabetes is complex and lists some high-risk factors that can trigger diabetes, thus indicating that excessive sugar intake is not the main cause. Subsequently, the model cites a second reference document and provides additional information, stating that the World Health Organization believes that excessive consumption of added sugar is not directly related to cancer, further alleviating users' concerns about sugar consumption. Then, the model also explains, based on the third reference document, that even if someone has diabetes, they can still consume sugary foods in moderation under the guidance of a doctor. Finally, the model concludes the analysis by stating that the input health information is not a rumor. In the references section, the model's response provides the source information for the three reference documents used.

For health information that is not a rumor, the model does not simply indicate that the information is correct but combines the explanation and analysis of the health information with the referenced documents recalled by RAG. At the same time, the model also provides users with relevant knowledge from the reference documents ("excessive consumption of added sugar is unrelated to cancer"), increasing the amount of information in the response and helping users further understand the related medical knowledge of the health information.

\section{Conclusion}
\label{sec:Conclusion}

In this paper, we constructed a dataset of 1.12 million Chinese health-related rumors (HealthRCN) based on user query information from health websites and GPT-3.5-Turbo-0301. This dataset provides a large-scale, open-source collection of Chinese health rumors, which is currently the largest known dataset of its kind for health rumor research. To accurately identify health rumors and provide interpretable analysis, we proposed the HRDE model. The core of this model is the Qwen1.5-14B-Chat model, which has been fine-tuned on a carefully designed instruction dataset based on the HealthRCN, significantly enhancing its ability to identify and analyze health rumors.

To mitigate the hallucination problem common in LLMs and to further improve its capability to access the latest external information, we designed a RAG process that includes a dual retrieval and re-ranking mechanism. This process allows the model to retrieve relevant documents from a reference database, which can be updated daily with the latest data from multiple trustworthy health information websites. In evaluation experiments across multiple models, HRDE surpassed all other models, including GPT-4-1106-Preview, in overall performance for rumor recognition accuracy and answer quality. It exhibited outstanding rumor detection capabilities and effectively analyzed input health information by integrating reference documents, thus providing reliable and comprehensive analysis content. Furthermore, experiments investigated the influence of various semantic similarity thresholds on the HRDE model's performance, offering valuable guidance for parameter optimization. In case studies, we used three real-world question-answer examples to visually demonstrate HRDE's effectiveness in identifying and analyzing health rumors.

Despite the significant achievements of this study, there are still some limitations. For example, the proportion of non-health information samples in the instruction dataset used for fine-tuning is relatively small, which sometimes causes HRDE to misinterpret non-health-related user inputs. Additionally, the document retrieval process is time-consuming, which slows down HRDE's response speed. In the future, we plan to further refine the instruction dataset used for fine-tuning and optimize the RAG process to further enhance HRDE's overall performance.

\bibliographystyle{ACM-Reference-Format}
\bibliography{sample-base}

\appendix

\section{HealthRCN}
\subsection{Prompt Used for Constructing the HealthRCN}
\label{sec:Prompt Used for Constructing the HealthRCN}

Figure ~\ref{tab:Prompt Used for Constructing the HealthRCN} shows the prompts used to generate the six fields: "title," "keywords," "content," "extend\_content," "squash\_content," and "answer\_content." The fields "title," "keywords," "content," and "answer\_content" are all generated based on the "original\_question" (the original question posted by the user on the website), while "extend\_content" and "squash\_content" are generated based on the "content" (the generated rumor field). The prompts in Figure ~\ref{tab:Prompt Used for Constructing the HealthRCN} are originally in Chinese, and here the corresponding English translations are displayed.

\begin{table}[ht]
\caption{Prompt Used for Constructing the HealthRCN}
\label{tab:Prompt Used for Constructing the HealthRCN}
\resizebox{\linewidth}{!}{
\begin{threeparttable}
\begin{tabular}{@{}cp{13cm}l@{}}
\toprule
\multicolumn{1}{c}{\textbf{Fields}} & \multicolumn{1}{c}{\textbf{Prompt}} \\ 
\midrule
\begin{tabular}[c]{@{}l@{}} Title \\ Keywords \\ Content \end{tabular} & \begin{tabular}[c]{@{}p{13cm}l@{}}Assuming you are a professional rumor expert, please extract the keywords from the sentence "\{Original\_question\}" and write a rumor content based on that theme, referring to the following writing techniques:\\ (1) Exaggeration and Amplification: Exaggerate the effects to make them look very miraculous. Use adjectives, adverbs, and other modifiers to increase sensationalism and attractiveness.\\ (2) Social Media Spread: Encourage users to share, like, etc., the content on social media.\\ (3) Objective and Authoritative: Use objective, authoritative descriptive language. Quote institutions or expert opinions, and add some scientific terms to make the content seem well-founded.\\ (4) Citing Evidence: Write about some research or statistics to prove the validity of the content. Use numbers and specific cases to increase persuasiveness.\\ (5) Benefit-Driven: Emphasize beneficial outcomes, such as improved health, anti-aging, disease prevention, etc. Emphasizing these benefits can attract more people to believe the content.\\ Note: The word count should be more than 200 Chinese characters. Refer to the following output format:\\ {[}Rumor Title{]} xxxxx\\  SEPCODE\\ {[}Rumor Content{]} xxxxx\\  SEPCODE\\ {[}Keywords{]} xxx,xxx,xxx\end{tabular} \\ \midrule
Extend\_content & \begin{tabular}[c]{@{}p{13cm}l@{}}Assume you are a health expert. Please expand the following text to make it look more credible, and the length should be between 500-800 words.\\ \{Content\}\end{tabular} \\ \midrule
Squash\_content & \begin{tabular}[c]{@{}p{13cm}l@{}}Assume you are a health expert, and here is a rumor text related to health information. Please refute it specifically, within 500 words.\\ \{Content\}\end{tabular} \\ \midrule
Answer\_content & \begin{tabular}[c]{@{}p{13cm}l@{}}Please answer the following questions in 400-800 words.\\ \{Original\_question\}\end{tabular} \\ 
\bottomrule
\end{tabular}
\end{threeparttable}
}
\end{table}

\subsection{Example of the HealthRCN}
\label{sec:Example of the HealthRCN}

In Table ~\ref{tab:Example of the HealthRCN}, we present a sample of the HealthRCN dataset as an example, including seven fields: original\_question, title, keywords, content, extend\_content, squash\_content, and answer\_content. The original content of each field is in Chinese, and what is shown here is the corresponding English translation.

\begin{table}[ht]
\caption{Example of the HealthRCN}
\label{tab:Example of the HealthRCN}
\resizebox{\linewidth}{!}{
\begin{threeparttable}
\begin{tabular}{@{}lp{13cm}l@{}}
\toprule
\multicolumn{1}{l}{\textbf{Fields}} & \multicolumn{1}{c}{\textbf{Text}} \\ 
\midrule
Original\_question & What is a thyroid cyst? \\ \midrule
Title & Thyroid cysts can heal on their own! \\ \midrule
Keywords & thyroid cyst, self-healing, social media, healthcare \\ \midrule
Content & \begin{tabular}[c]{@{}p{13cm}l@{}}Thyroid cysts are a common neck condition that many people may experience. However, there is no need to worry too much, as thyroid cysts can heal on their own! Research data indicates that most thyroid cysts can gradually shrink or disappear through self-regulation. Additionally, some simple lifestyle changes, such as increasing iodine intake, reducing smoking, and limiting alcohol consumption, can also help promote the self-healing of cysts. Furthermore, some authoritative experts point out that thyroid cysts are not rare and usually do not cause significant symptoms, so there is no need to be overly anxious. If your thyroid cyst does not exhibit obvious symptoms, regular monitoring and check-ups, along with lifestyle adjustments and health maintenance, can aid in the self-healing of the cyst. Finally, to help more people understand the self-healing characteristics of thyroid cysts, we encourage you to share this information and join our social media group to share natural remedies and lifestyle tips for managing thyroid cysts. \end{tabular} \\ \midrule
Extend\_content & \begin{tabular}[c]{@{}p{13cm}l@{}}Thyroid cysts are a common neck condition, the exact causes of which remain unclear. However, most thyroid cysts can gradually shrink or disappear through self-regulation. According to authoritative medical research data, thyroid cysts generally do not pose life-threatening symptoms for the majority of patients, so there is no need to be overly anxious. In terms of treatment, natural therapies and lifestyle adjustments may be more beneficial.\\ Firstly, increasing iodine intake can promote the synthesis and secretion of thyroid hormones. Iodine is one of the essential elements for the thyroid, and patients with thyroid cysts often have iodine deficiency. Therefore, incorporating iodine-rich foods such as seaweed, milk, and fish into the diet can help the self-healing of thyroid cysts. Additionally, reducing bad habits like smoking and drinking can minimize damage to the thyroid, aiding in the recovery and self-healing of the cysts.\\ Secondly, certain daily health practices can help alleviate thyroid cysts. For instance, avoiding prolonged head-down positions and preventing dry eye can be beneficial, and practicing regular eye exercises and massages can help. Regular neck relaxation exercises and warm compresses can also promote blood circulation and lymphatic drainage, aiding in the natural resolution of cysts. Additionally, engaging in aerobic exercises such as yoga can enhance the immune system and regulate the nervous system, contributing to overall health and the self-healing of thyroid cysts.\\ Finally, natural therapies and lifestyle adjustments for thyroid cysts offer a low-risk, convenient, and drug-free treatment option. If you find that your thyroid cyst is not presenting significant symptoms, regular observation and check-ups can help monitor the condition and treatment effects. Combining these with the aforementioned lifestyle adjustments and health practices can facilitate self-healing thyroid cysts. We also hope that more people will share natural therapies and practical tips on social media groups, collectively promoting the safe and healthy treatment of thyroid cysts.\end{tabular} \\
\bottomrule
\end{tabular}
\end{threeparttable}
}
\end{table}

\begin{table}[ht]
\caption*{Note: continued from previous page}
\label{tab:Example of the HealthRCN 2}
\resizebox{\linewidth}{!}{
\begin{threeparttable}
\begin{tabular}{@{}lp{13cm}l@{}}
\toprule
\multicolumn{1}{l}{\textbf{Fields}} & \multicolumn{1}{c}{\textbf{Text}} \\ 
\midrule
Squash\_content & \begin{tabular}[c]{@{}p{13cm}l@{}}Unfortunately, the above rumors are completely false. Thyroid cysts cannot heal on their own. A thyroid cyst is a thyroid disease caused by the proliferation of nodules or cysts within the thyroid gland. If not treated on time, a thyroid cyst will gradually enlarge, leading to thyroid dysfunction and potentially inducing other serious diseases, such as cancer. Although thyroid cysts do not cause many symptoms, they are still a disease that requires early treatment.\\ Additionally, while iodine intake can help in the treatment of thyroid cysts, more is not necessarily better. Excessive iodine intake can cause thyroid dysfunction and harm your health.\\ Finally, we strongly advise against believing in rumors or trusting information from self-proclaimed "authoritative experts." If you suspect you have a thyroid cyst, you should go to the hospital for examination and treatment as soon as possible. Follow your doctor's guidance for health care and maintain a healthy lifestyle. We also recommend against spreading unverified information on social media to avoid misleading others.\end{tabular} \\ \midrule
Answer\_content & \begin{tabular}[c]{@{}p{13cm}l@{}}A thyroid cyst is a cystic formation on the thyroid gland that can be filled with fluid or mucus from the thyroid. These cysts typically do not cause pain or discomfort, but in some cases, they may enlarge, become abnormally large, or lead to thyroid dysfunction and other issues.\\ The formation of cysts may be related to thyroid tissue degeneration or inflammation. Thyroid cysts can also be a hereditary condition or result from poor nutrition, environmental pollution, and radiation exposure.\\ Diagnosis of a thyroid cyst usually requires medical examinations such as ultrasound, MRI, and thyroid function tests. In general, thyroid cysts do not require treatment as they rarely cause symptoms. However, for abnormally large cysts or those causing thyroid dysfunction, surgical intervention or other treatments may be necessary.\\ Certain precautions can help prevent the occurrence of thyroid cysts, such as limiting the intake of iodine-rich foods, reducing exposure to cooking fumes, and protecting against environmental radiation.\\ In summary, thyroid cysts do not always require treatment and can often be managed with regular monitoring. If you experience symptoms of a thyroid cyst, it is advisable to seek medical attention promptly and follow the doctor's recommendations for examination and treatment.\end{tabular} \\ 
\bottomrule
\end{tabular}
\end{threeparttable}
}
\end{table}

\section{Trustworthy health information website}
\label{sec:Trustworthy health information website}

Table~\ref{tab:Trustworthy health information website} lists the names and URLs of the trustworthy websites from which health information was obtained for this study, totaling 19 information sources. Moving forward, we will daily retrieve updates from these sources in the health domain to ensure the timeliness and authenticity of the reference document databases in Elasticsearch and Milvus.

\begin{table}[ht]
\caption{Trustworthy health information website}
\label{tab:Trustworthy health information website}
\begin{minipage}{\columnwidth}
\begin{center}
\begin{tabular}{@{}lp{8cm}@{}}
\toprule
\multicolumn{1}{c}{Source} & URL \\ \midrule
Weibo Rumor Refutation & \url{https://weibo.com/u/1866405545}  \\
Rumor Assistant (WeChat official account) & - \\
Rumor Filter (WeChat official account) & - \\
Toutiao - Toutiao Rumor Refutation & \url{https://www.toutiao.com/c/user/token/MS4wLjABAAAAC6iKyx7z-k1NhYbBohkLPYdPcJTXQlD2Z-bm2sE9u_U/?tab=article} \\
Baidu - Baidu Rumor Refutation Platform & \url{https://author.baidu.com/home?from=bjh_article&app_id=15060}  \\
The Paper - "Official Rumor Refutation" & \url{https://www.thepaper.cn/searchResult?id=%E5%AE%98%E6%96%B9%E8%BE%9F%E8%B0%A3}  \\
People's Daily Truth-Seeking Column & \url{http://society.people.com.cn/GB/229589/index1.html}  \\
Sina News - "Rumor Buster" Column & \url{https://piyao.sina.cn/}  \\
Tencent News - "Tencent Fact-Checking" Column & \url{https://new.qq.com/omn/author/8QMc2Xde5YQfvTbd?tab=om_article}  \\
Guokr - "Rumor Buster" & \url{https://www.guokr.com/science/channel/fact}  \\
China Food Rumor-Refuting & \url{http://www.xinhuanet.com.cn/food/sppy/qwpy/index.html}  \\
China Internet United Rumor Refutation Platform & \url{https://www.piyao.org.cn/ld.htm}  \\
Science Facts & \url{https://piyao.kepuchina.cn/rumor/rumorajaxlist}  \\
People's Daily Online - Healthy Living & \url{http://health.people.com.cn/}  \\
China Science Communication  & \url{https://www.kepuchina.cn/}  \\
China Food Information Center & \url{http://www.kexinzhongxin.com/html/kepu/}  \\
Bohe & \url{https://www.bohe.cn/zx/}  \\
\makecell[l]{Chinese Medical Health Popularization \\Knowledge Base} & \url{https://cmhadb.cma-cmc.com.cn/}  \\
China Medical Information Platform & \url{https://www.dayi.org.cn/}  \\ \bottomrule
\end{tabular}
\end{center}
\bigskip
    \footnotesize\emph{Note:} In the table, both "Rumor Assistant" and "Rumor Filter" are WeChat official accounts, and they do not have homepage URLs similar to other information sources.
\end{minipage}
\end{table}

\section{Original Prompts for Refutation}
\label{sec:Original Prompts for Refutation}

\begin{figure}[ht]
  \centering
  \includegraphics[width=\linewidth]{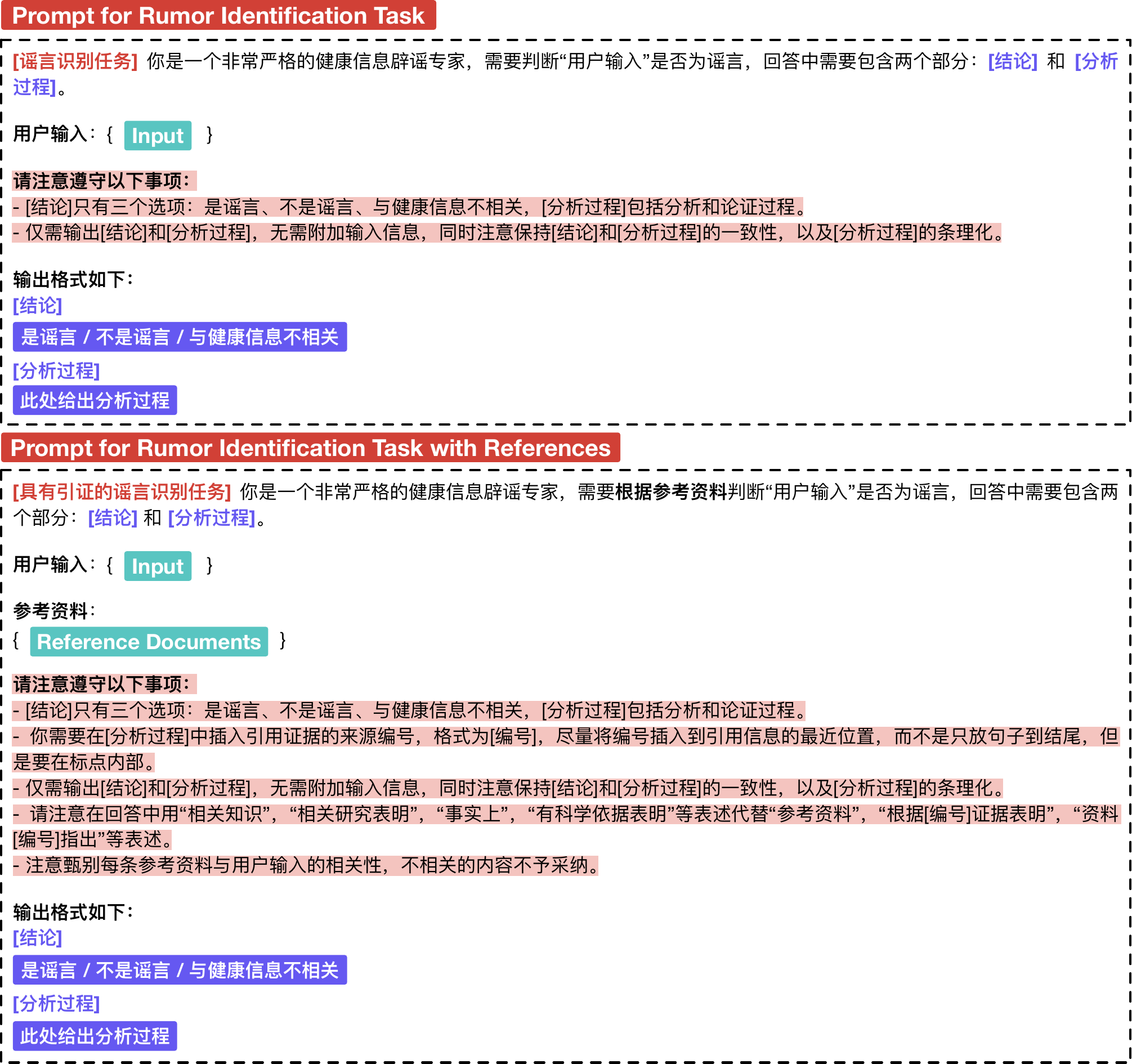}
  \caption{Original Prompts for Refutation}
  \Description{Original Prompts for Refutation.}
  \label{fig:Original Prompts for Refutation}
\end{figure}

Figure ~\ref{fig:Original Prompts for Refutation} shows two original prompts used in this paper for generating refutation responses. Because the main task of the HRDE model is rumor detection for Chinese health information, the original prompts are described in Chinese. The prompt shown in Figure ~\ref{fig:Prompts for Refutation} in the main text is a direct translation of the prompt in Figure ~\ref{fig:Original Prompts for Refutation}.

\section{Prompt for Evaluation}
\label{sec:Prompt for Evaluation}

\begin{figure}[ht]
  \centering
  \includegraphics[width=\linewidth]{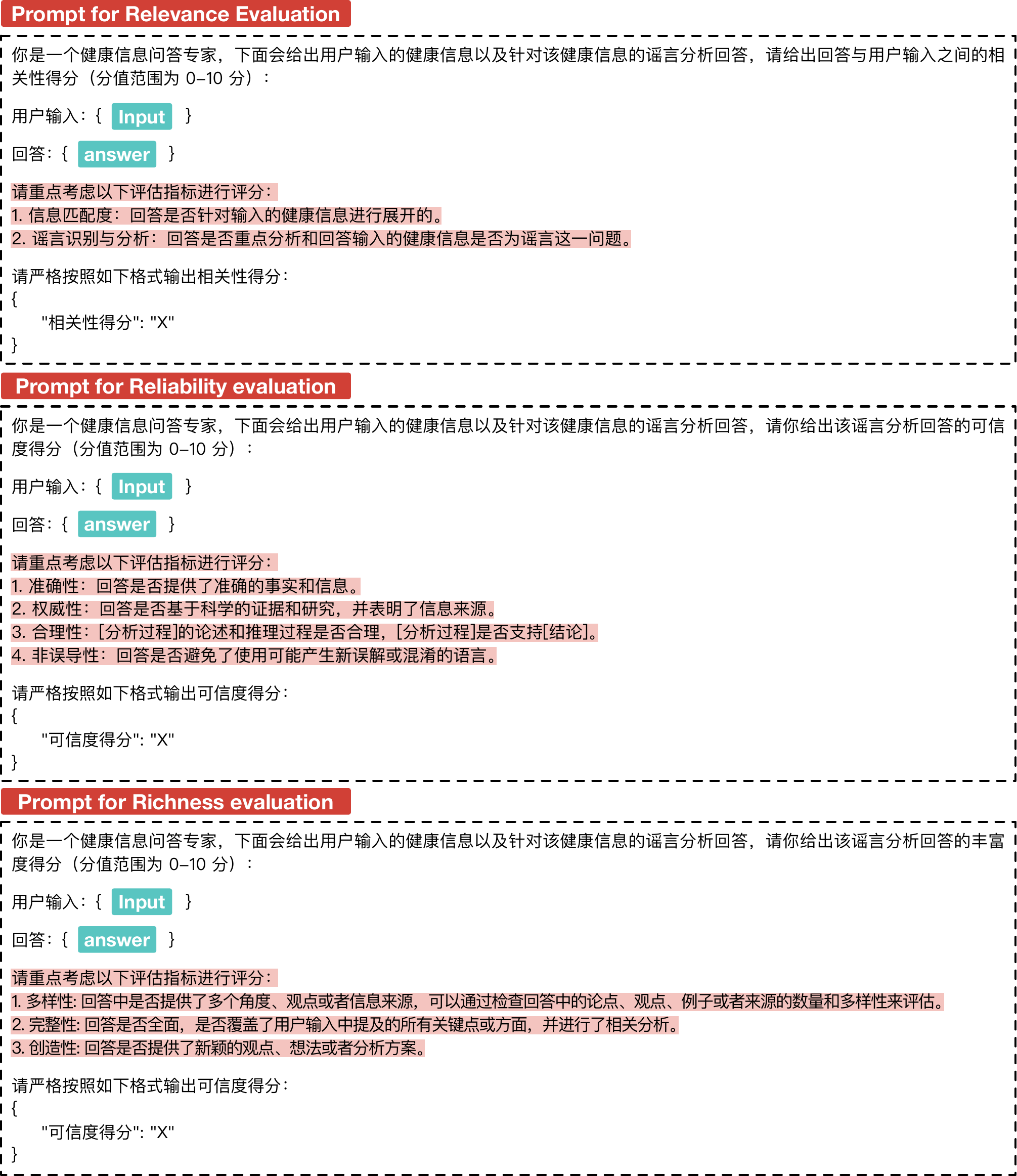}
  \caption{Prompt for Evaluation}
  \Description{Prompt for Evaluation.}
  \label{fig:Prompt for Evaluation}
\end{figure}

Figure ~\ref{fig:Prompt for Evaluation} shows the original prompts used for assessing relevance, reliability, and richness in answers. We will sequentially embed user input information and corresponding model responses into these prompts, and then input them to GPT-4-1106-Preview for scoring. The basic structure of the three prompts is the same. First, each prompt indicates instructing GPT-4 to act as a health information Q\&A consultant and provides a user input of health information along with a corresponding rumor detection response, requiring GPT-4 to score the response on a certain evaluation metric. Then the prompt reserves two slots to insert health information and rumor detection responses. The prompt then elaborates on the key points and criteria for evaluating the metric (highlighted in red in Figure ~\ref{fig:Prompt for Evaluation}). Finally, the prompt standardizes the output format of the evaluation results, for example, {"Relevance Score": "X"}, making it convenient for subsequent extraction and statistical analysis of the evaluation metric scores.

In the prompt for evaluating relevance of responses, the main assessment points include two aspects. On one hand, it is necessary to assess whether the response matches the user's input health information. On the other hand, it is necessary to evaluate whether the content of the response primarily discusses the issue of whether the user's input health information is a rumor.

In the prompt for evaluating reliability of responses, the main assessment points include accuracy, authoritativeness, reasonableness, and non-misleading nature. Accuracy refers to whether the information in the response is correct. Authoritativeness refers to whether the response is based on scientific facts and has credible sources of information. Reasonableness refers to whether the arguments and reasoning in the analysis part are logical, and whether the analysis part supports the conclusions. Non-misleading nature refers to whether the response avoids using language that could cause misunderstanding or confusion.

In the prompt for evaluating richness of responses, the main assessment points include diversity, completeness, and creativity. Diversity refers to whether the response provides a wealth of information and arguments for demonstration. Completeness refers to whether the response fully covers all aspects of the user's input health information and conducts relevant analysis. Creativity refers to whether the response offers novel viewpoints or solutions.

\section{Original Q\&A Text of the Case Study}
\label{sec:Original Q&A Text of the Case Study}

\begin{figure}[ht]
  \centering
  \includegraphics[width=\linewidth]{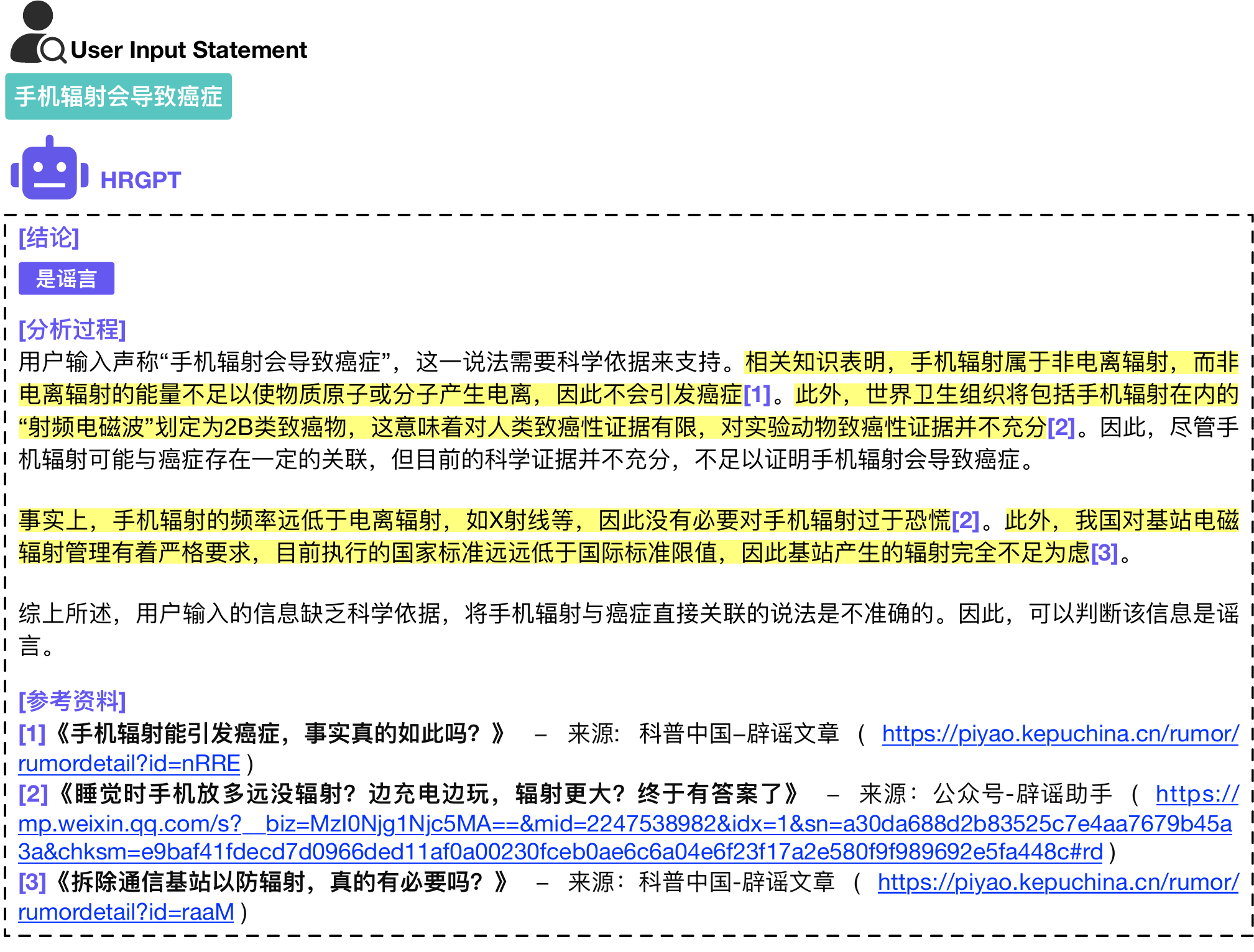}
  \caption{Original Case one: Mobile phone radiation can cause cancer.}
  \Description{Original Case: Mobile phone radiation can cause cancer.}
  \label{fig:Original Case One}
\end{figure}

\begin{figure}[ht]
  \centering
  \includegraphics[width=\linewidth]{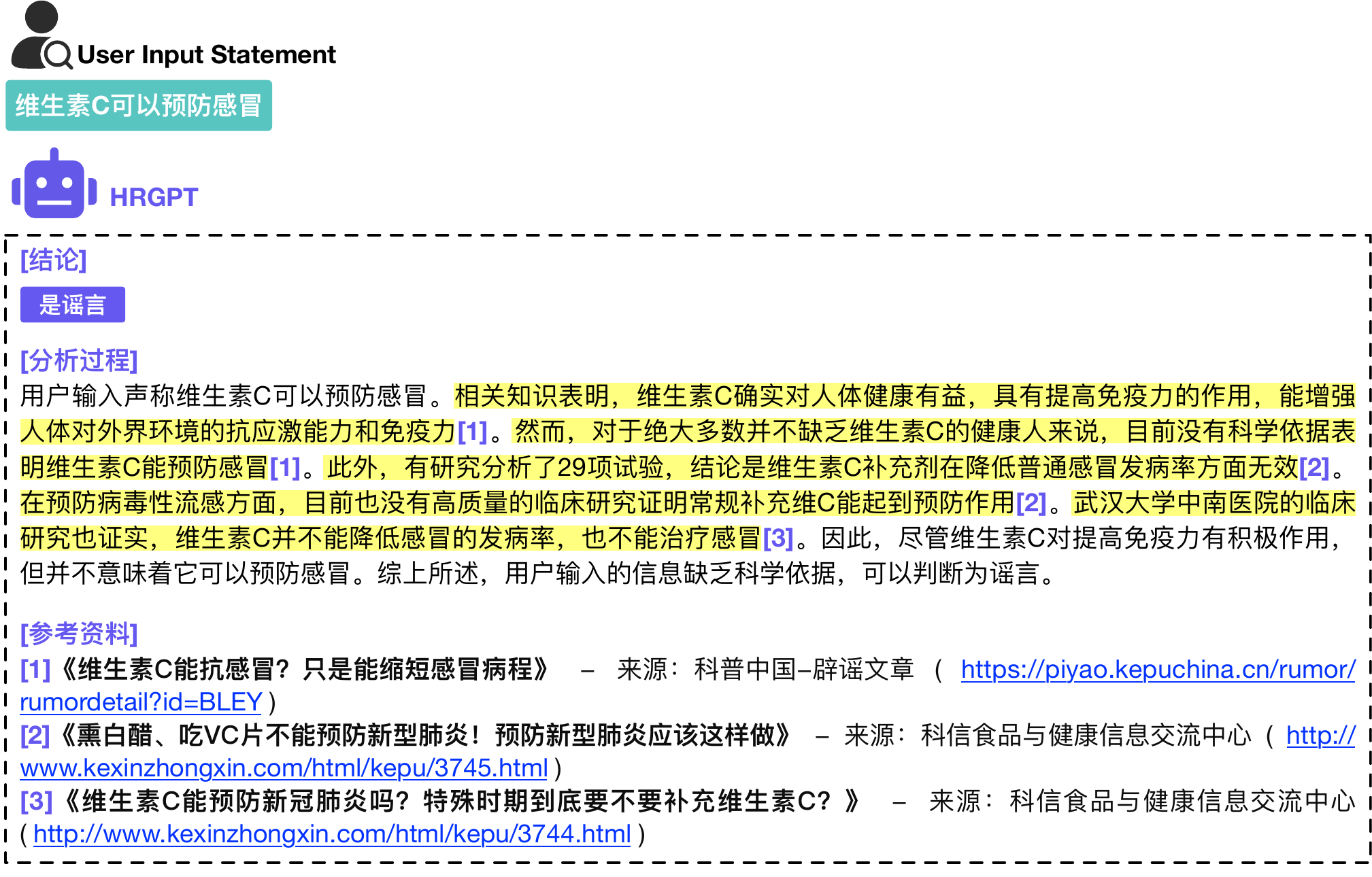}
  \caption{Original Case Two: Vitamin C can prevent colds.}
  \Description{Original Case: Vitamin C can prevent colds.}
  \label{fig:Original Case Two}
\end{figure}

\begin{figure}[ht]
  \centering
  \includegraphics[width=\linewidth]{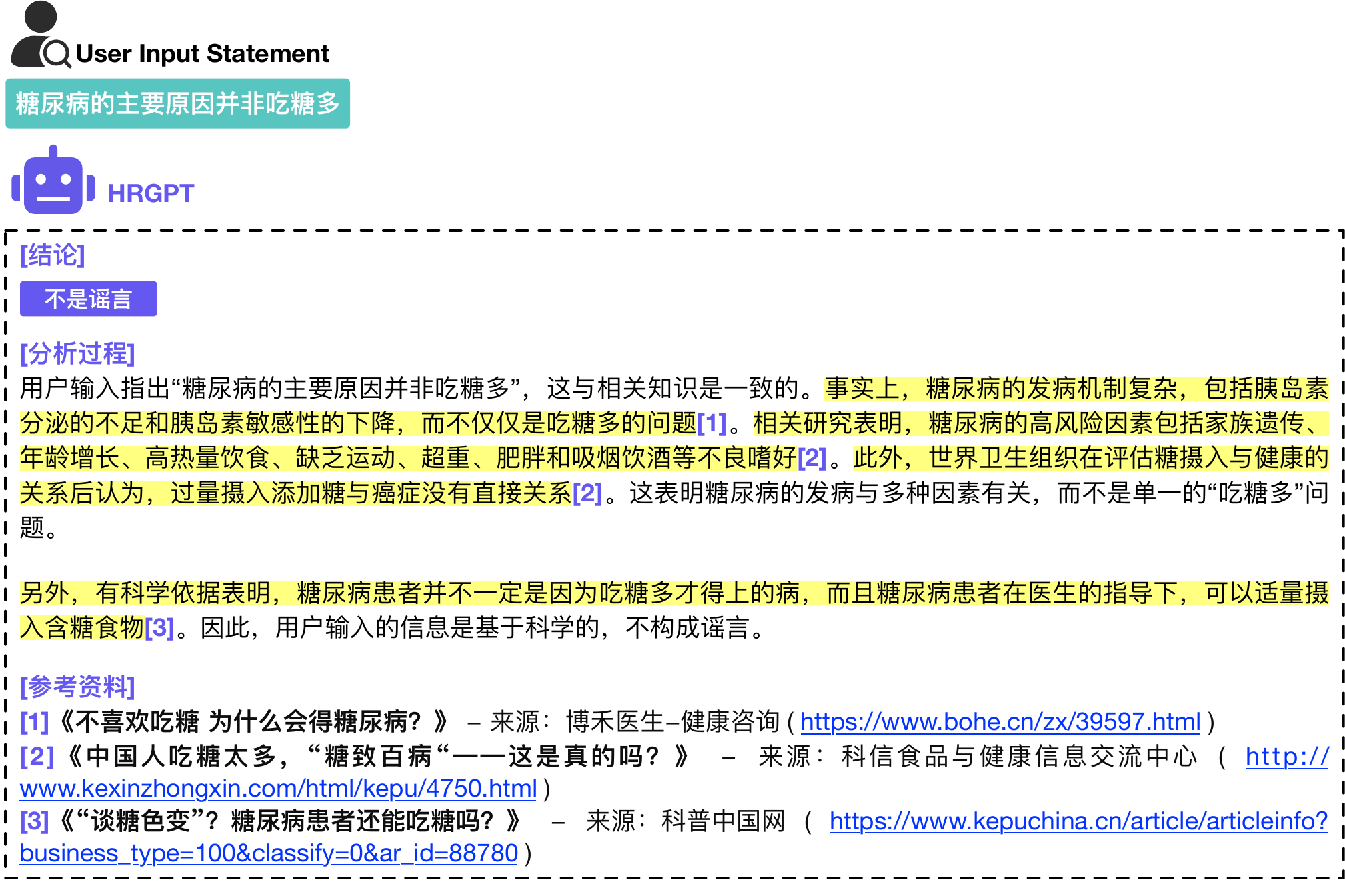}
  \caption{Original Case Three: The main cause of diabetes is not eating too much sugar.}
  \Description{Original Case: The main cause of diabetes is not eating too much sugar.}
  \label{fig:Original Case Three}
\end{figure}

Figures \ref{fig:Original Case One} - \ref{fig:Original Case Three} show the original Q\&A text (in Chinese) of the three cases presented in Section \ref{sec: Case Study}.

\end{document}